\documentclass[10pt]{article}
\usepackage[margin=0.9in]{geometry} 

\usepackage[utf8]{inputenc}
\usepackage[T1]{fontenc}
\usepackage{times}
\usepackage{amsmath,amsfonts,amssymb,amsthm}
\usepackage{graphicx}
\usepackage{xcolor}
\usepackage{booktabs}
\usepackage{multirow}
\usepackage{caption}
\usepackage{subcaption}
\usepackage{tikz}
\usepackage{pgfplots}
\usepackage{algorithm}
\usepackage{algorithmic}
\usepackage{listings}
\usepackage{fancyvrb}
\usepackage{mdframed}
\usepackage{hyperref}
\usepackage[numbers,sort&compress]{natbib}

\usepackage{balance} 
\usepackage{appendix}
\usepackage{enumitem} 
\usepackage{array}
\usepackage{textcomp} 
\usepackage{lipsum} 

\usetikzlibrary{shapes,arrows,positioning,fit,calc,patterns,decorations.pathreplacing}
\pgfplotsset{compat=1.17} 

\definecolor{deepblue}{rgb}{0,0,0.5}
\definecolor{deepred}{rgb}{0.6,0,0}
\definecolor{deepgreen}{rgb}{0,0.5,0}
\definecolor{lightgray}{rgb}{0.95,0.95,0.95}
\definecolor{midgray}{rgb}{0.85,0.85,0.85}
\definecolor{darkgray}{rgb}{0.4,0.4,0.4}

\lstdefinestyle{python}{
    language=Python,
    basicstyle=\ttfamily\footnotesize,
    keywordstyle=\color{deepblue}\bfseries,
    commentstyle=\color{deepgreen}\itshape,
    stringstyle=\color{deepred},
    numbers=left,
    numberstyle=\tiny\color{darkgray},
    numbersep=5pt,
    backgroundcolor=\color{lightgray},
    frame=single,
    framerule=0pt,
    breaklines=true,
    breakatwhitespace=false,
    showspaces=false,
    showstringspaces=false,
    showtabs=false,
    tabsize=2,
    captionpos=b,
    morekeywords={self,True,False,None}
}

\theoremstyle{plain}
\newtheorem{theorem}{Theorem}[section] 

\theoremstyle{definition}
\newtheorem{definition}[theorem]{Definition}

\newtheorem{remark}[theorem]{Remark}

\newcommand{\IP}{\texttt{IP}}
\newcommand{\IA}{\texttt{IA}}
\newcommand{\STRG}{\texttt{STRG}}
\newcommand{\NW}{\texttt{NW}}
\newcommand{\SOI}{\texttt{SOI}}
\newcommand{\VR}{\texttt{VR}}

\hypersetup{
    colorlinks=true,
    linkcolor=deepblue,
    citecolor=deepgreen,
    urlcolor=deepred,
    pdftitle={Cognitive Weave: A Dynamic Insight Synthesis System for Advanced Agent Memory},
    pdfauthor={AI Research Collective},
    pdfsubject={Research Paper on AI Agent Memory},
    pdfkeywords={AI, LLM, Agent Memory, Cognitive Systems, Knowledge Representation, Insight Synthesis}
}

\title{\textbf{Cognitive Weave: Synthesizing Abstracted Knowledge with a Spatio-Temporal Resonance Graph}}

\author{
  Akash Vishwakarma \\
  \textit{University of Southern} California \\
  \texttt{vishwaka@usc.edu}
  \and
  Hojin Lee \\
  \textit{WorkOnward, Inc} \\
  \texttt{hojin@workonward.com}
  \and
  Mohith Suresh \\
  \textit{University of Southern} California \\
  \texttt{mohiths@usc.edu}
  \and
  Priyam Shankar Sharma \\
  \textit{Yeshiva University} \\
  \texttt{psharma2@mail.yu.edu}
  \and
  Rahul Vishwakarma \\
  \textit{IEEE Senior Member} \\
  \texttt{rahul.vishwakarma@ieee.org}
  \and
  Sparsh Gupta \\
  \textit{University of Southern California} \\
  \texttt{sparshg@usc.edu}
  \and
  Yuvraj Anupam Chauhan \\
  \textit{Northeastern University} \\
  \texttt{chauhan.yuv@northeastern.edu}
}

\date{}

\begin{document}

\maketitle

\begin{abstract}
The emergence of capable Large Language Model (LLM) based agents necessitates memory architectures that transcend mere data storage, enabling continuous learning, nuanced reasoning, and dynamic adaptation. Current memory systems often grapple with fundamental limitations in structural flexibility, temporal awareness, and the ability to synthesize higher-level insights from raw interaction data. This paper introduces \textbf{Cognitive Weave}, a novel memory framework centered around a multi-layered Spatio-Temporal Resonance Graph (\STRG). This graph manages information as semantically rich Insight Particles (\IP{}s), which are dynamically enriched with Resonance Keys, Signifiers, and Situational Imprints via a dedicated Semantic Oracle Interface (\SOI). These \IP{}s are interconnected through typed Relational Strands, forming an evolving knowledge tapestry. A key of Cognitive Weave is the Cognitive Refinement process, an autonomous mechanism that includes the synthesis of Insight Aggregates (\IA{}s)—condensed, higher-level knowledge structures derived from identified clusters of related \IP{}s. We present comprehensive experimental results demonstrating Cognitive Weave's marked enhancement over existing approaches in long-horizon planning tasks, evolving question-answering scenarios, and multi-session dialogue coherence. The system achieves a notable 34\% average improvement in task completion rates and a 42\% reduction in mean query latency when compared to state-of-the-art baselines. Furthermore, this paper also explores the ethical considerations inherent in such advanced memory systems, discusses the implications for long-term memory in LLMs, and outlines promising future research trajectories.
\end{abstract}

\setcounter{page}{1} 

\section{Introduction}
\label{sec:introduction}

The remarkable advancements in Large Language Models (LLMs) \cite{Vaswani2017, Brown2020, Achiam2023} have spurred the development of increasingly autonomous agents. These agents are designed to perform complex tasks and engage in sophisticated interactions within dynamic and often unpredictable environments \cite{Park2023, Wang2023agent, Xi2023}. A common architectural pattern involves augmenting LLMs with external tools, memory systems, and structured workflows to bolster their innate reasoning, planning, and execution capabilities \cite{Yao2022, Shinn2023}. Among these augmentations, the memory system stands as a critical, yet often underdeveloped, component that underpins an agent's capacity for long-term interaction, continuous learning, and genuine adaptation. Without robust memory, agents remain predominantly reactive, unable to build upon past experiences or develop a persistent understanding of their world.

Initial solutions for equipping LLMs with memory predominantly focused on Retrieval Augmented Generation (RAG) \cite{Lewis2020, Karpukhin2020, Borgeaud2022}. RAG architectures provide LLMs with access to historical information or domain knowledge, typically through vector-based similarity searches over static or slowly changing knowledge bases. While RAG has proven effective for enhancing factual grounding and reducing hallucinations in knowledge-intensive NLP tasks, the requirements of long-term autonomy, continuous experiential learning, and the synthesis of novel insights reveal significant limitations in these foundational approaches. Many contemporary agent memory systems struggle with:

\begin{itemize}
    \item {Structural Rigidity and Lack of Flexibility:} Many systems rely on predefined schemas or simplistic data structures that do not readily adapt to the diverse and evolving nature of information encountered by an agent across various tasks and contexts \cite{Packer2023, Lee2024}. This rigidity can hinder the representation of complex, nuanced relationships and impede the integration of novel types of information.
    \item {Deficient Temporal Awareness and Reasoning:} The ability to understand and reason about the temporal aspects of information—such as the order of events, the duration of states, or the evolution of entities over time—is often inadequately supported \cite{Maharana2024}. Simple timestamping is insufficient for nuanced temporal reasoning critical for many real-world applications.
    \item {Inability to Perform Insight Synthesis and Abstraction:} Most memory systems function as passive repositories, lacking the intrinsic capability to autonomously synthesize higher-level understanding, abstractions, or novel insights from the accumulated raw data \cite{Xu2025, Sun2024}. Agents are thus limited in their ability to generalize from specific instances or form complex conceptual models.
\end{itemize}
To address these profound challenges, we propose \textbf{Cognitive Weave}, a novel memory architecture designed to function as an active, evolving tapestry of interconnected insights rather than a static datastore. Cognitive Weave reconceptualizes agent memory as a dynamic cognitive substrate. The system's core innovations are engineered to directly tackle the aforementioned limitations:

\begin{itemize}
    \item A sophisticated hybrid Spatio-Temporal Resonance Graph (\STRG) which integrates vectorial, temporal, and rich relational data layers to provide a flexible and multifaceted representation of knowledge.

    \item Insight Particles (\IP{}s) and Insight Aggregates (\IA{}s) as the primary, semantically rich units of memory, allowing for both granular information storage and the representation of synthesized, higher-order knowledge.

    \item A Semantic Oracle Interface (\SOI), leveraging advanced LLMs for deep semantic processing, interpretation, and structuring of incoming information into \IP{}s and for synthesizing \IA{}s.

    \item A dynamic Cognitive Refinement process that enables continuous learning, knowledge synthesis, importance recalibration, and structural evolution of the memory graph.
\end{itemize}

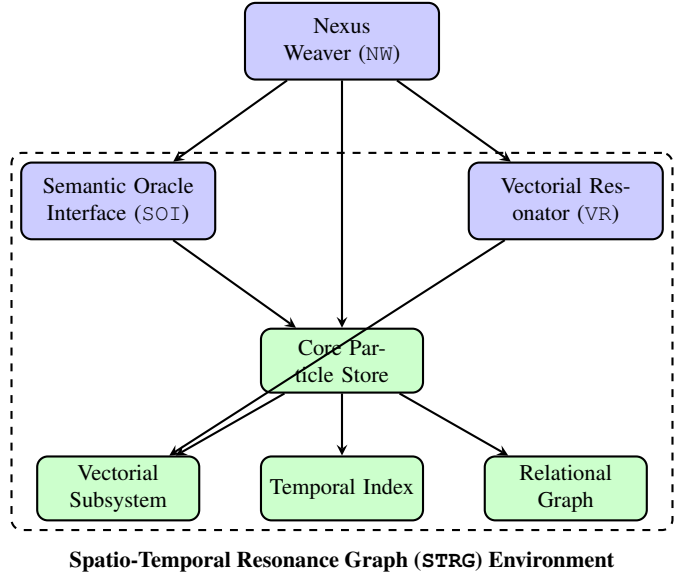
\begin{figure}[htbp]
\centering
\begin{tikzpicture}[scale=0.85, transform shape] 
    \tikzstyle{component} = [rectangle, rounded corners, minimum width=3cm, minimum height=1.2cm, text width=2.8cm, align=center, draw=black, fill=blue!20, thick]
    \tikzstyle{layer} = [rectangle, rounded corners, minimum width=2.5cm, minimum height=1cm, text width=2.3cm, align=center, draw=black, fill=green!20, thick]
    \tikzstyle{arrow} = [thick,->,>=stealth]
    
    \node[component] (nw) at (0,0) {Nexus Weaver (\NW)};
    \node[component] (soi) at (-3.5,-2.5) {Semantic Oracle Interface (\SOI)};
    \node[component] (vr) at (3.5,-2.5) {Vectorial Resonator (\VR)};
    
    \node[layer] (core) at (0,-5) {Core Particle Store};
    \node[layer] (vector) at (-3.5,-7) {Vectorial Subsystem};
    \node[layer] (temporal) at (0,-7) {Temporal Index};
    \node[layer] (relational) at (3.5,-7) {Relational Graph};
    
    \draw[arrow] (nw) -- (soi);
    \draw[arrow] (nw) -- (vr);
    \draw[arrow] (nw) -- (core); 
    
    \draw[arrow] (soi) -- (core);
    \draw[arrow] (vr) -- (vector);
    
    \draw[arrow] (core) -- (vector);
    \draw[arrow] (core) -- (temporal);
    \draw[arrow] (core) -- (relational);
    
    \node[draw, dashed, thick, rounded corners, fit=(core) (vector) (temporal) (relational) (soi) (vr), label={[yshift=-0.2cm]south:\textbf{Spatio-Temporal Resonance Graph (\STRG) Environment}}] (strg_boundary) {};
\end{tikzpicture}
\caption{High-level architecture of the Cognitive Weave system, illustrating the interplay between its principal components and the layered architecture of the Spatio-Temporal Resonance Graph (\STRG). The Nexus Weaver (\NW) orchestrates the Semantic Oracle Interface (\SOI) and Vectorial Resonator (\VR), which process information for storage and interaction within the multi-layered \STRG. The \STRG{} itself comprises the Core Particle Store, Vectorial Subsystem, Temporal Index, and Relational Graph layers. This paper elaborates on the theoretical foundations, architectural design, and empirical validation of Cognitive Weave, demonstrating its potential to significantly advance the state of agent memory systems.}
\label{fig:architecture}
\end{figure}

\section{Related Work}
\label{sec:related_work}

The pursuit of effective memory systems for AI agents is not new, and research in this area has progressed through several distinct paradigms. Each wave of innovation has sought to address the shortcomings of its predecessors, gradually moving towards more dynamic, structured, and contextually aware memory architectures. Understanding this evolution provides essential context for appreciating the design choices and contributions of Cognitive Weave.

\subsection{Early Memory Models and Retrieval Augmented Generation}
Early approaches often involved simple logging of interaction histories or curated knowledge bases. The advent of LLMs led to the widespread adoption of Retrieval Augmented Generation (RAG) \cite{Lewis2020, Karpukhin2020}. RAG systems typically employ vector databases populated with embeddings of text chunks. When an LLM needs to generate a response or perform a task, relevant information is retrieved based on semantic similarity (cosine similarity between query and document embeddings) and provided as augmented context to the LLM \cite{Borgeaud2022}. While RAG significantly improves factual grounding and access to external knowledge, standard implementations often treat the knowledge base as relatively static and lack mechanisms for dynamic learning, complex relationship modeling, or temporal reasoning beyond simple recency.

\subsection{Context Window Management and Hierarchical Memory}
A significant challenge for LLMs is their finite context window. Systems like MemGPT \cite{Packer2023} conceptualize LLM memory as an operating system, employing a virtual memory management technique with paging to move information between the limited physical context (the LLM's direct input) and external storage. This allows LLMs to effectively manage much larger histories than their native context windows would permit, enhancing capabilities in long-running conversations and tasks requiring extended context. MemoryBank \cite{Zhong2024} further explored dynamic memory updating, drawing inspiration from human forgetting curves to manage the relevance of stored information. While these systems represent important advances in extending effective memory capacity and managing information flow, they may still be constrained by relatively simple underlying data structures or predefined organizational schemas, potentially limiting their adaptability across highly diverse or rapidly evolving informational domains \cite{Lee2024}. Cognitive Weave aims to build upon these concepts by providing a more intrinsically flexible and semantically rich storage and synthesis fabric.

\subsection{Graph-Based and Structured Memory Architectures}
Recognizing the limitations of flat or purely vectorial memory, researchers have increasingly turned to graph-based structures for more flexible and expressive knowledge organization. Graph databases allow for the explicit representation of entities and their relationships, enabling more complex queries and reasoning pathways.

Mem0~\cite{Mem0_2025} is a recent example that combines graph structures with RAG principles, aiming to create a more organized memory layer for AI agents. Similarly, A-MEM \cite{Xu2025} proposes an "agentic memory" system that creates interconnected knowledge networks through dynamic indexing and linking, drawing inspiration from the Zettelkasten method \cite{Kadavy2021}. The Zettelkasten approach, originating from note-taking methodologies, emphasizes the creation of atomic, richly interconnected notes to foster emergent understanding. This philosophy of interconnected atomic units of knowledge is a strong influence on Cognitive Weave's Relational Strands concept, which links \IP{}s.
Cognitive Weave extends these graph-based approaches by integrating not only relational data but also explicit temporal and vectorial layers within a unified \STRG{} framework, coupled with a formal process for synthesizing higher-order \IA{}s.

\subsection{Temporal and Spatio-Temporal Reasoning in Memory}
While many systems incorporate timestamps or apply recency biases \cite{Packer2023}, dedicated mechanisms for sophisticated temporal reasoning have been less explored in mainstream agent memory architectures. The ability to understand sequences, durations, and the evolution of information over time is crucial for many real-world tasks. Zep's Graphiti engine \cite{ZepBlog2024} is an example of a system moving towards temporal knowledge graphs, highlighting the growing recognition of this need.
The STMS patent by Hawkins et al. \cite{Hawkins2013} outlined concepts for spatio-temporal pattern encoding and prediction, forming part of the theoretical motivation for Cognitive Weave's integrated Spatio-Temporal Resonance Graph. While STMS focuses on pattern recognition in sensory data streams, its core idea of memory encoding spatio-temporal relationships influenced our desire to build a memory system where time is a first-class citizen, not merely metadata. Cognitive Weave's Temporal Index Layer and the temporal metadata within \IP{}s aim to provide robust support for such reasoning.

\subsection{LLM Memory Architectures (Long Contexts)}
A major challenge for large language models (LLMs) is handling long or ongoing contexts because of fixed input length and the quadratic cost of self-attention.  
Recent work has explored architectures that either extend effective context length or incorporate external memory.  
For example, MemLong augments a decoder-only LLM with a non-differentiable memory retriever and fine-grained retrieval–attention; this allows the model to offload long-term context to an external store and fetch relevant history on the fly, extending the usable context from approximately 4\,k tokens to 80\,k on a single GPU \citep{liu2024memlong}.  
Such memory-augmented LLMs outperform standard long-context baselines on downstream benchmarks.  
Focused Transformer improves long-context utilization via contrastive training that teaches the model to ignore irrelevant tokens, enabling LongLLaMA variants with context windows up to 256\,k \citep{tworkowski2023focused}.  
Unlike these methods, which modify the transformer or attention pattern, Cognitive Weave avoids exorbitant context lengths by dynamically storing and retrieving only salient information.  
Rather than solely extending raw context, our system emphasises insight synthesis—compressing and structuring past knowledge into high-level memories.  
This distinction lets Cognitive Weave maintain long-term coherence without the full quadratic cost, complementing approaches like MemLong by focusing on what to remember (insights) in addition to how long to remember it.

\subsection{Retrieval-Augmented Generation and Structured Memory}
Retrieval-Augmented Generation (RAG) stores documents in a vector database and injects top-ranked chunks into the prompt but relies on coarse semantic similarity and treats each chunk in isolation, often missing cross-chunk relationships or introducing irrelevant context \citep{Lewis2020}.  
Structured memory representations have emerged to address these limitations.  
TOBUGraph builds a knowledge graph from unstructured text on the fly using an LLM to extract entities and relations; queries are answered by graph traversal rather than flat embedding lookup, yielding significantly higher precision and recall than RAG in personal-memory benchmarks \citep{kashmira2024graph}.  
By eliminating arbitrary text chunking, TOBUGraph reduces information fragmentation and hallucination.  
Cognitive Weave shares the goal of richer memory representation but weaves distributed insights and contextual links into a dynamic memory store instead of constructing an explicit knowledge graph.  
Thus both systems go beyond flat vector memory; Cognitive Weave focuses on creating new, abstracted knowledge (insight synthesis), whereas TOBUGraph explicitly formalises knowledge into graph nodes.

\subsection{Memory-Augmented Agents and Dynamic Recall}
There is growing interest in equipping autonomous LLM-based agents with long-term memory that evolves over time.  
A recent survey highlights memory as essential for complex, long-horizon tasks and catalogues mechanisms used in LLM agents \citep{zhang2024survey}.  
For conversational assistants, MemInsight continuously augments and indexes interaction history; intelligent summarisation and embedding improved recommendation persuasiveness by 14 \% and boosted retrieval recall by 34 \% over a RAG baseline in dialogue tasks \citep{salama2025meminsight}.  
Hou \textit{et al.}\ integrate human-like recall and consolidation: each past event receives a recall probability that decays over time, and only sufficiently memorable events are recalled; distilled long-term memories are stored with timestamps to enable temporal context \citep{hou2024my}.  
Cognitive Weave aligns with these trends by dynamically curating memories; like MemInsight, we maintain semantically rich indexes, and like Hou \textit{et al.}, we prioritise and fuse memories in a human-like way.  
However, Cognitive Weave additionally synthesises higher-level insights from raw history, allowing the agent to recall broader implications rather than isolated facts.

\subsection{Task-Oriented (Workflow) Memory}
LLM-based agents in decision-making environments benefit from remembering and reusing successful strategies.  
Agent Workflow Memory (AWM) induces reusable action routines from past tasks and injects them into future planning, substantially improving success rates on long-horizon web-navigation benchmarks \citep{wang2024agent}.  
AriGraph combines a knowledge-graph world model with episodic memory in simulated environments; the agent’s memory graph supports associative retrieval of facts relevant to the current goal and achieves zero-shot completion of complex TextWorld games \citep{anokhin2024arigraph}.  
These works underscore the value of structured and contextualised memory.  
Cognitive Weave complements them by dynamically updating its knowledge fabric with each new insight, enabling the agent not only to remember events but also to reason over their synthesized interconnections across episodes.

\subsection{Synthesis of Capabilities}
Table \ref{tab:comparison} provides a comparative summary of the capabilities of several representative memory system paradigms, highlighting how Cognitive Weave aims to synthesize and extend features found disparately in prior work. The table underscores Cognitive Weave's ambition to deliver a comprehensive solution that integrates vectorial retrieval, graph-based structuring, temporal awareness, dynamic evolution, and, crucially, insight synthesis within a single, coherent architecture.

\begin{table}[htbp]
\centering
\caption{Comparative overview of memory system capabilities. Cognitive Weave aims to integrate all listed capabilities within a unified framework, moving beyond specialized solutions.}
\label{tab:comparison}
\resizebox{\textwidth}{!}{%
\begin{tabular}{@{}lccccc@{}}
\toprule
\textbf{System / Paradigm} & \textbf{Vectorial Retrieval} & \textbf{Graph Structure} & \textbf{Temporal Awareness} & \textbf{Insight Synthesis} & \textbf{Dynamic Evolution} \\
\midrule
Standard RAG & \checkmark &  &  &  &  \\
MemGPT \cite{Packer2023} & \checkmark &  & \checkmark (Implicit/Context Mgt) &  & \checkmark (Memory Mgt) \\
Mem0 \cite{Mem0_2025} & \checkmark & \checkmark &  &  & \checkmark \\
A-MEM \cite{Xu2025} & \checkmark & \checkmark &  & \checkmark (Evolution/Linking) & \checkmark \\
Zep Graphiti \cite{ZepBlog2024} & \checkmark & \checkmark & \checkmark (Explicit TKG) &  & \checkmark \\
\textbf{Cognitive Weave (Proposed)} & \checkmark & \checkmark & \checkmark (Explicit Layer) & \checkmark (IA Synthesis) & \checkmark (Refinement) \\
\bottomrule
\end{tabular}%
}
\end{table}

Cognitive Weave thus positions itself not as an incremental improvement in one specific area, but as a holistic advancement aiming to weave together these disparate strengths into a more powerful and integrated cognitive memory architecture for next-generation AI agents.

\section{The Cognitive Weave System Architecture}
\label{sec:system}

Cognitive Weave introduces a paradigm where agent memory is not a passive archive but an active, evolving cognitive substrate. It is engineered to transform raw experiential data into a rich, interconnected tapestry of insights that dynamically adapt and grow with the agent. This section provides a detailed exposition of its core architectural components and the foundational concepts that govern its operation.

\begin{figure*}[htbp]
    \centering
    \includegraphics[width=\linewidth]{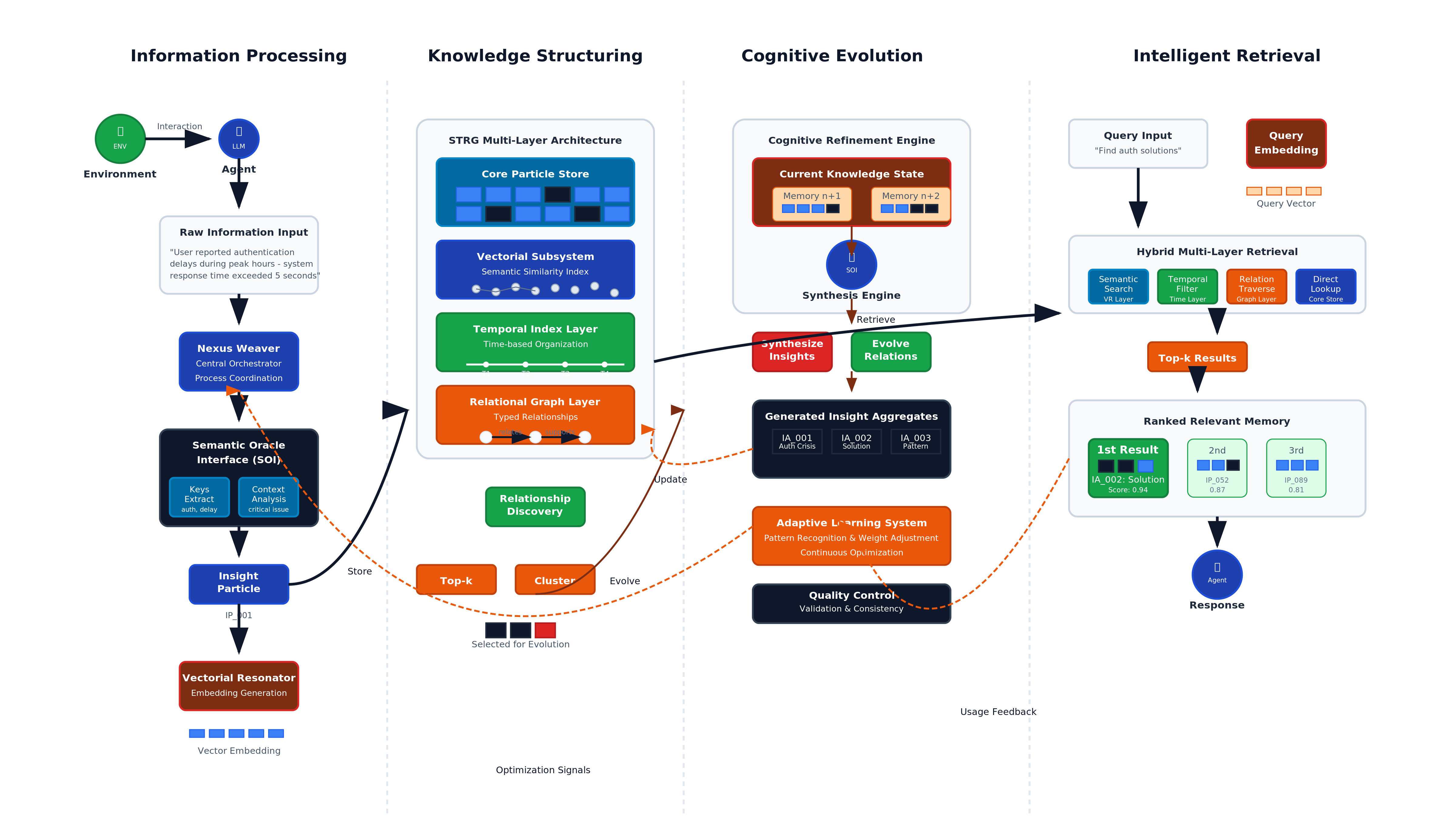}
    \caption{The Cognitive Weave System Architecture}
    \label{fig:cognitive_weave}
\end{figure*}

\subsection{Core Architectural Components}
The Cognitive Weave system is composed of four primary, synergistic components, as depicted in Figure \ref{fig:architecture}. These components collaborate to manage the lifecycle of information, from ingestion and processing to storage, refinement, and retrieval.

\subsubsection{Nexus Weaver (\NW)}
The Nexus Weaver (\NW) serves as the central nervous system and orchestrator of the Cognitive Weave architecture. It is responsible for managing the complete lifecycle of Insight Particles (\IP{}s), from their initial creation to their eventual archival or pruning. The \NW{} initiates and oversees the Cognitive Refinement cycles, which are crucial for the evolution of the memory tapestry. Furthermore, it handles all incoming recall requests, determining the optimal strategy to retrieve relevant information from the Spatio-Temporal Resonance Graph (\STRG). Key responsibilities of the \NW{} include:
\begin{itemize}
    \item \textbf{Information Flow Coordination:} Directing the flow of data between the Semantic Oracle Interface (\SOI), the Vectorial Resonator (\VR), and the various layers of the \STRG{}. This ensures that information is processed, embedded, and stored in a consistent and efficient manner.
    \item \textbf{Refinement Process Management:} Triggering Cognitive Refinement processes based on a set of configurable conditions. These conditions might include temporal triggers (e.g., periodic refinement), event-based triggers (e.g., ingestion of a significant new piece of information), resource availability, or metrics indicating memory saturation or fragmentation.
    \item \textbf{Integrity and Concurrency Management:} Maintaining the structural and semantic integrity of the \STRG{}. This involves managing concurrent access requests (reads, writes, updates) and ensuring consistency across the distributed components of the memory system, potentially employing transactional semantics or optimistic locking mechanisms where appropriate.
    \item \textbf{Query Processing and Dispatch:} Receiving recall requests, interpreting query intent (possibly with LLM assistance), and dispatching optimized sub-queries to the relevant \STRG{} layers (e.g., vector search, graph traversal, temporal filtering).
\end{itemize}
The \NW{} is envisioned as an intelligent component, potentially incorporating its own learning mechanisms to optimize its orchestration strategies over time.

\subsubsection{Semantic Oracle Interface (\SOI)}
The Semantic Oracle Interface (\SOI) is the primary engine for deep semantic understanding and structuring within Cognitive Weave. It is powered by advanced Large Language Models (LLMs), such as those available through Azure OpenAI services (e.g., o4-mini \cite{AzureO4Mini2024} or more powerful equivalents), selected for their robust natural language understanding, generation, and reasoning capabilities. The \SOI{} performs a critical transformation function, converting raw, unstructured or semi-structured input data into highly structured Insight Particles (\IP{}s). This transformation is formalized as:
\begin{equation}
\text{\SOI}_{\text{transform}}: \text{RawData} \rightarrow \{\text{ResonanceKeys}, \text{Signifiers}, \text{SituationalImprint}, \text{CoreData}_{\text{structured}}\}
\label{eq:soi_transform}
\end{equation}
where $\text{RawData}$ can be text, dialogue turns, observations, or other forms of agent experience. The output includes:
\begin{itemize}
    \item \textbf{Resonance Keys ($\mathcal{K}$):} A set of keywords, topics, or conceptual tags that capture the essence of the data, facilitating faceted search and associative retrieval.
    \item \textbf{Signifiers ($\mathcal{S}$):} Semantic markers indicating the type, nature, or intent of the information (e.g., assertion, question, observation, hypothesis).
    \item \textbf{Situational Imprint ($\mathcal{M}$):} Contextual metadata describing the circumstances surrounding the information's acquisition (e.g., source, agent's internal state, environmental conditions).
    \item \textbf{Core Data ($\mathcal{D}$):} The primary content of the information, potentially cleaned, summarized, or canonicalized by the \SOI{}.
\end{itemize}
Beyond initial \IP{} creation, the \SOI{} plays a pivotal role in the Cognitive Refinement process, specifically in the synthesis of Insight Aggregates (\IA{}s) from clusters of related \IP{}s, as detailed in Section \ref{sec:cognitive_refinement}.

\subsubsection{Vectorial Resonator (\VR)}
The Vectorial Resonator (\VR) is responsible for translating the textual or semantic content of Insight Particles (\IP{}s) into dense vector embeddings. These embeddings capture the semantic essence of the \IP{}s in a high-dimensional vector space, enabling efficient similarity-based search and clustering. The \VR{} employs state-of-the-art sentence embedding models, such as Sentence-BERT \cite{Reimers2019} or similar transformer-based architectures optimized for producing semantically meaningful representations. The embedding function can be represented as:
\begin{equation}
\text{\VR}_{\text{embed}}: \mathcal{D}_{\text{IP}} \rightarrow \mathbf{v} \in \mathbb{R}^d
\label{eq:vr_embed}
\end{equation}
where $\mathcal{D}_{\text{IP}}$ is the core data content of an \IP{}, and $\mathbf{v}$ is a $d$-dimensional real-valued vector. The dimensionality $d$ is typically determined by the chosen embedding model (e.g., 768 or 1024 for many BERT-based models). These embeddings are then managed by the Vectorial Resonance Subsystem within the \STRG{} (Section \ref{sec:vectorial_subsystem}) for fast nearest neighbor searches.

\subsubsection{Spatio-Temporal Resonance Graph (\STRG)}
The Spatio-Temporal Resonance Graph (\STRG{}) is the heart of Cognitive Weave's memory storage. It is not a monolithic database but a hybrid, multi-layered data structure designed to capture the multifaceted nature of information. Its architecture, detailed further in Section \ref{sec:strg_details}, integrates distinct layers for storing core particle data, managing vector embeddings, indexing temporal information, and representing rich relational connections. This composite structure allows Cognitive Weave to support diverse query types and reasoning processes that go beyond the capabilities of simpler memory models. The "Spatio-Temporal" aspect refers to its ability to represent information situated in both a conceptual "space" (via relational and vectorial connections) and along a temporal dimension. "Resonance" alludes to the system's ability to activate and retrieve relevant patterns of information based on incoming queries or new data, akin to resonance in physical systems.

\subsection{Insight Particles (\IP{}s): The Atomic Units of Memory}
Central to the Cognitive Weave paradigm is the concept of the Insight Particle (\IP{}). An \IP{} is not merely a raw piece of data but a meticulously structured knowledge unit, designed to be the atomic constituent of the agent's memory. Each \IP{} encapsulates not only the core informational content but also a rich set of metadata that contextualizes it and facilitates its integration into the broader knowledge tapestry.

\begin{definition}[Insight Particle (\IP{})]
\label{def:ip}
An Insight Particle (\IP{}) is formally defined as a tuple $\mathcal{I} = \langle id, \mathcal{D}, \mathcal{K}, \mathcal{S}, \mathcal{M}, \mathcal{T}, \mathcal{R}, \mathcal{A} \rangle$, where:
\begin{itemize}
    \item $id$: A unique identifier for the \IP{}, ensuring addressability and referential integrity.
    \item $\mathcal{D}$: The core data content of the particle. This could be a textual snippet, a structured observation, a dialogue excerpt, or other forms of agent experience. The \SOI{} may process raw input into a canonical or summarized form for $\mathcal{D}$.
    \item $\mathcal{K}$: A set of Resonance Keys. These are keywords, topics, or conceptual tags extracted or inferred by the \SOI{}, designed to capture the thematic essence of $\mathcal{D}$ and facilitate associative retrieval.
    \item $\mathcal{S}$: A set of Signifiers. These are semantic markers assigned by the \SOI{} that denote the nature, type, or pragmatic function of the information (e.g., factual assertion, hypothesis, query, emotional state).
    \item $\mathcal{M}$: The Situational Imprint. This is a collection of metadata describing the context in which the information was acquired or generated, such as data source, agent's internal state at the time, environmental parameters, or task relevance.
    \item $\mathcal{T}$: Temporal Metadata. This includes timestamps for creation ($t_{\text{create}}$), last modification ($t_{\text{modify}}$), and last access ($t_{\text{access}}$), as well as potentially a valid time range or event time associated with the content of $\mathcal{D}$.
    \item $\mathcal{R}$: A set of Relational Strands. These represent typed, directed connections to other \IP{}s or \IA{}s within the \STRG{}, forming the relational fabric of the memory (e.g., `supports`, `contradicts`, `elaborates`, `causes`).
    \item $\mathcal{A}$: Access and Importance Metrics. This includes attributes like access frequency ($f_{\text{access}}$), a dynamically computed importance score ($I$), and potentially other metrics related to utility or decay.
\end{itemize}
\end{definition}

This multi-faceted representation allows \IP{}s to be indexed, queried, and reasoned over through various dimensions—semantic similarity of $\mathcal{D}$ (via its vector embedding), keyword matching on $\mathcal{K}$, filtering by $\mathcal{S}$ or $\mathcal{M}$, temporal range queries on $\mathcal{T}$, and graph traversal via $\mathcal{R}$. This richness is key to enabling nuanced organization and retrieval that surpasses the capabilities of systems relying solely on vector similarity or simple key-value storage. The creation and ongoing enrichment of \IP{}s are orchestrated by the \NW{}, primarily leveraging the \SOI{} for semantic content and the \VR{} for vectorial representation.

\subsection{Spatio-Temporal Resonance Graph (\STRG{}) Layered Architecture}
\label{sec:strg_details}
The \STRG{} is the cornerstone of Cognitive Weave's memory, providing a sophisticated, multi-layered infrastructure for storing, organizing, and retrieving \IP{}s and \IA{}s. It is designed as a hybrid system that synergistically combines the strengths of different data management paradigms.

\subsubsection{Core Particle Store}
The foundation of the \STRG{} is the Core Particle Store. This layer is responsible for the persistent storage of the full \IP{} and \IA{} objects, including all their structured attributes as defined in Definition \ref{def:ip}. It is envisioned to be implemented using a flexible NoSQL database, such as MongoDB or a similar document-oriented database. The choice of a NoSQL solution offers several advantages:
\begin{itemize}
    \item \textbf{Schema Flexibility:} The structure of \IP{}s or \IA{}s may evolve as the agent learns or as new types of information are encountered. NoSQL databases readily accommodate such schema evolution without requiring costly migrations.
    \item \textbf{Rich Data Structures:} Document databases are well-suited for storing complex, nested objects like \IP{}s with their diverse set of attributes.
    \item \textbf{Scalability:} Many NoSQL systems are designed for horizontal scalability, which is crucial for handling the potentially vast amounts\_of memory an agent might accumulate over its lifetime.
    \item \textbf{Efficient Operations:} These databases often provide efficient indexing on multiple attributes and support for bulk read/write operations, facilitating rapid ingestion and retrieval of particles.
\end{itemize}
The Core Particle Store ensures data integrity and durability, potentially supporting ACID-like properties for critical updates, depending on the specific NoSQL system chosen and its configuration.

\subsubsection{Vectorial Resonance Subsystem}
\label{sec:vectorial_subsystem}
To enable rapid semantic similarity search, the Vectorial Resonance Subsystem manages the dense vector embeddings of \IP{}s (and potentially \IA{}s) generated by the \VR{}. This subsystem implements highly efficient Approximate Nearest Neighbor (ANN) search algorithms. A key technology underpinning this layer is FAISS (Facebook AI Similarity Search) \cite{Johnson2019}, a library renowned for its performance in billion-scale similarity searches. The primary operation supported is finding the $k$ \IP{}s whose embeddings $\mathbf{v}_i$ are most similar to a given query embedding $\mathbf{v}_q$, typically measured by cosine similarity:
\begin{equation}
\text{sim}(\mathbf{v}_i, \mathbf{v}_q) = \frac{\mathbf{v}_i \cdot \mathbf{v}_q}{\|\mathbf{v}_i\| \|\mathbf{v}_q\|}
\label{eq:cosine_similarity}
\end{equation}
This subsystem allows the agent to quickly retrieve memories that are semantically related to a current query or observation, even if they do not share exact keywords. It works in concert with the Core Particle Store, storing embeddings that correspond to the full \IP{} data.

\subsubsection{Temporal Index Layer}
Addressing the critical need for temporal awareness, the Temporal Index Layer maintains specialized indices on the temporal metadata ($\mathcal{T}$) of \IP{}s. This typically involves creating B-tree or similar sorted indices on attributes such as creation time ($t_{\text{create}}$), modification time ($t_{\text{modify}}$), and last access time ($t_{\text{access}}$). These indices allow for:
\begin{equation}
\mathcal{T}_{\text{index}} = \text{IndexOn}(\{t_{\text{create}}, t_{\text{modify}}, t_{\text{access}}, t_{\text{event\_start}}, t_{\text{event\_end}}\})
\label{eq:temporal_index}
\end{equation}
The inclusion of $t_{\text{event\_start}}$ and $t_{\text{event\_end}}$ allows for indexing based on the actual time the content of the \IP{} refers to, not just its metadata lifecycle. This layer enables efficient execution of temporal range queries (e.g., "retrieve all \IP{}s created last week" or "find observations related to event X between time A and B") with logarithmic time complexity, $O(\log N)$, where $N$ is the number of indexed particles. This capability is fundamental for tasks requiring historical context, trend analysis, or reasoning about sequences of events.

\subsubsection{Relational Strand Graph Layer}
The Relational Strand Graph Layer explicitly models the relationships ($\mathcal{R}$) between \IP{}s and \IA{}s, forming a rich, interconnected knowledge graph. In this layer:
\begin{itemize}
    \item \textbf{Nodes} represent individual Insight Particles (\IP{}s) or Insight Aggregates (\IA{}s).
    \item \textbf{Edges} (Relational Strands) represent typed, directed relationships between these nodes.
\end{itemize}
The types of relationships are crucial for capturing nuanced semantic connections. Cognitive Weave proposes a vocabulary of edge types, which can be extended, including:
\begin{itemize}
    \item \texttt{supports}: Indicates that one \IP{} provides evidence for another.
    \item \texttt{contradicts}: Indicates that one \IP{} presents conflicting information to another.
    \item \texttt{elaborates}: Signifies that one \IP{} provides more detail or explanation for another.
    \item \texttt{causes}: Represents a causal link between the content of two \IP{}s.
    \item \texttt{precedes}: Denotes a temporal ordering if not captured by event times.
    \item \texttt{derivedFrom}: Links an \IA{} to the constituent \IP{}s from which it was synthesized.
\end{itemize}
This graph structure, illustrated abstractly in Figure \ref{fig:relational_graph}, enables complex reasoning through graph traversal, pattern detection (e.g., finding chains of supporting evidence or identifying contradictory clusters), and understanding the provenance of synthesized knowledge. The creation and maintenance of these strands are key functions of the Cognitive Refinement process, often guided by the \SOI{}.

\begin{figure}[htbp]
\centering
\begin{tikzpicture}[scale=0.9, transform shape] 
    \tikzstyle{ipnode} = [circle, minimum width=1.2cm, draw=black, fill=blue!30, thick, align=center]
    \tikzstyle{ianode} = [ellipse, minimum width=1.5cm, minimum height=1cm, draw=black, fill=red!30, thick, align=center]
    \tikzstyle{relation} = [thick,->,>=stealth]
    
    \node[ipnode] (ip1) at (0,0) {\IP{}1};
    \node[ipnode] (ip2) at (2.5,1.5) {\IP{}2};
    \node[ipnode] (ip3) at (2.5,-1.5) {\IP{}3};
    \node[ipnode] (ip4) at (5,0) {\IP{}4};
    \node[ianode] (ia1) at (7.5,0) {\IA{}1};
    
    \draw[relation] (ip1) to [bend left=15] node[above, midway, sloped, font=\tiny] {supports} (ip2);
    \draw[relation] (ip1) to [bend right=15] node[below, midway, sloped, font=\tiny] {elaborates} (ip3);
    \draw[relation] (ip2) to [bend left=15] node[above, midway, sloped, font=\tiny] {causes} (ip4);
    \draw[relation] (ip3) to [bend right=15] node[below, midway, sloped, font=\tiny] {relatedTo} (ip4); 
    
    \draw[relation, dashed, blue!70] (ip2) to [bend left=10] node[above, midway, sloped, font=\tiny] {derivedFrom} (ia1);
    \draw[relation, dashed, blue!70] (ip3) to node[pos=0.4, right, font=\tiny] {derivedFrom} (ia1);
    \draw[relation, dashed, blue!70] (ip4) to [bend right=10] node[below, midway, sloped, font=\tiny] {derivedFrom} (ia1);

    \node at (3.75, -3) [text width=8cm, align=center] {\small An illustrative fragment of the Relational Strand Graph showing \IP{}s linked by typed relationships. An \IA{} (IA1) is synthesized from a cluster of related \IP{}s (IP2, IP3, IP4).};
\end{tikzpicture}
\caption{Conceptual example of a Relational Strand Graph fragment within Cognitive Weave. Insight Particles (\IP{}s) are interconnected by typed relationships. An Insight Aggregate (\IA{}) is shown as being synthesized from, and thus related to, a cluster of constituent \IP{}s. This graph structure facilitates nuanced reasoning and knowledge discovery.}
\label{fig:relational_graph}
\end{figure}

\subsection{Cognitive Refinement Process}
\label{sec:cognitive_refinement}
A defining characteristic of Cognitive Weave is its dynamic nature, embodied by the Cognitive Refinement process. This is an ongoing, autonomous set of operations orchestrated by the \NW{} that continuously evolves and improves the memory tapestry. It transforms the memory from a passive store into an active learning system. This process comprises three principal mechanisms: Insight Aggregate Synthesis, Relational Strand Management, and Importance Recalibration.

\subsubsection{Insight Aggregate (\IA{}) Synthesis}
The synthesis of Insight Aggregates (\IA{}s) is perhaps the most innovative aspect of Cognitive Refinement. \IA{}s are condensed, higher-level knowledge structures that represent emergent understanding derived from clusters of related \IP{}s. They serve to abstract information, identify salient patterns, and create conceptual shortcuts for more efficient reasoning and retrieval. The synthesis process, outlined in Algorithm \ref{alg:ia_synthesis}, involves several key steps:

\begin{algorithm}[htbp]
\caption{Insight Aggregate (\IA{}) Synthesis Process}
\label{alg:ia_synthesis}
\begin{algorithmic}[1]
\STATE \textbf{Input:} Current state of the Spatio-Temporal Resonance Graph (\STRG{}), clustering threshold $\tau_{\text{cluster}}$, synthesis trigger condition $\Omega_{\text{synth}}$
\STATE \textbf{Output:} Set of new Insight Aggregates ($\mathcal{S}_{\text{IA}}$) integrated into \STRG{}
\WHILE{$\Omega_{\text{synth}}$ is met}
    \STATE $\mathcal{C}_{\text{clusters}} \leftarrow$ IdentifyPotentialClusters(\STRG{}, $\tau_{\text{cluster}}$)
    \COMMENT{Clusters based on semantic similarity, relational proximity, temporal coherence}
    \FOR{each identified cluster $c \in \mathcal{C}_{\text{clusters}}$}
        \IF{ClusterQualityCheck($c$) passes}
            \STATE $\text{Data}_{\text{cluster}} \leftarrow$ ExtractConsolidatedData($c$)
            \COMMENT{Gather core data and key metadata from \IP{}s in $c$}
            \STATE $\text{Prompt}_{\text{synth}} \leftarrow$ GenerateSynthesisPrompt($\text{Data}_{\text{cluster}}$)
            \STATE $\mathcal{I}_{\text{new\_IA}} \leftarrow$ \SOI.SynthesizeInsight($\text{Prompt}_{\text{synth}}$)
            \COMMENT{\SOI{} (LLM) generates the new \IA{}}
            \STATE AddToSTRG($\mathcal{I}_{\text{new\_IA}}$)
            \COMMENT{Store the new \IA{} in Core Particle Store, create its vector embedding, temporal indices}
            \STATE EstablishRelations($\mathcal{I}_{\text{new\_IA}}$, $c$, type=\texttt{derivedFrom})
            \COMMENT{Link the new \IA{} to its constituent \IP{}s}
            \STATE $\mathcal{S}_{\text{IA}} \leftarrow \mathcal{S}_{\text{IA}} \cup \{\mathcal{I}_{\text{new\_IA}}\}$
        \ENDIF
    \ENDFOR
\ENDWHILE
\RETURN $\mathcal{S}_{\text{IA}}$
\end{algorithmic}
\end{algorithm}

The `IdentifyPotentialClusters` step (Line 4) is a crucial heuristic that may combine multiple criteria: semantic similarity of \IP{} embeddings (from the Vectorial Resonance Subsystem), connectivity in the Relational Strand Graph, and temporal proximity or co-occurrence (from the Temporal Index Layer). The `ClusterQualityCheck` (Line 6) ensures that only meaningful and coherent clusters are processed for synthesis. The actual synthesis (Line 9) is performed by the \SOI{}, which receives the consolidated data from the cluster and a carefully crafted prompt to generate a concise yet comprehensive summary or abstraction that becomes the new \IA{}. This \IA{} is then integrated into the \STRG{} like any other particle, including having its own relational strands, notably `derivedFrom` links to its source \IP{}s. This process allows Cognitive Weave to "learn" by creating new, abstracted knowledge from existing information.

\subsubsection{Relational Strand Management}
The relational fabric of the \STRG{} is not static; it evolves through dynamic Relational Strand Management. This involves several sub-processes:
\begin{itemize}
    \item \textbf{Pattern-Based Strand Suggestion and Creation:} The system, potentially guided by the \SOI{} or learned heuristics within the \NW{}, can identify patterns among \IP{}s (e.g., frequent co-occurrence in similar contexts, strong semantic relationships between unlinked particles) and suggest or automatically create new Relational Strands. For instance, if multiple \IP{}s describing prerequisites are consistently followed by an \IP{} describing an outcome, a `causes` or `precedes` strand might be inferred.
    \item \textbf{Confidence Score Adjustment:} Relational Strands can have associated confidence scores, reflecting the system's certainty about the validity or strength of the relationship. These scores can be dynamically adjusted based on new incoming information, user feedback (if applicable), or consistency checks with other parts of the graph.
    \item \textbf{Contradiction Detection and Resolution (Preliminary):} A highly advanced and challenging aspect is the detection of contradictory information within the \STRG{}. Cognitive Weave aims to identify \IP{}s or clusters of \IP{}s that present conflicting assertions. Initial resolution strategies might involve flagging contradictions, reducing the importance of conflicting \IP{}s, or creating special \IA{}s that explicitly represent the contradiction and its sources. Robust contradiction resolution is a long-term research goal (see Section \ref{sec:limitations}).
\end{itemize}
This ongoing management ensures the Relational Strand Graph remains a relevant and accurate representation of the agent's understanding.

\subsubsection{Importance Recalibration}
Not all memories hold equal importance or relevance over time. Cognitive Weave implements an Importance Recalibration mechanism to dynamically adjust the significance of \IP{}s and \IA{}s. The importance score $I$ of a particle is updated based on factors such as access frequency, its role in successful task completion, its connections to highly important \IA{}s, and temporal decay. A generalized form of this update, extending the one in the original text, could be:
\begin{equation}
I_{\text{new}} = f(I_{\text{old}}, \lambda_{\text{decay}}, f_{\text{access}}, C_{\text{task}}, \sum w_k \cdot I_{\text{linked\_IA}_k}, \text{UserFeedback})
\label{eq:importance_recalibration}
\end{equation}
where $I_{\text{old}}$ is the previous importance, $\lambda_{\text{decay}}$ is a decay factor (see Section \ref{sec:temporal_decay_model}), $f_{\text{access}}$ is access frequency, $C_{\text{task}}$ represents contribution to task success, $\sum w_k \cdot I_{\text{linked\_IA}_k}$ reflects weighted importance derived from linked \IA{}s, and UserFeedback (if available) provides explicit relevance signals. The original paper's formula, $I_{\text{new}} = \alpha \cdot I_{\text{old}} + \beta \cdot f_{\text{access}} + \gamma \cdot \text{IA}_{\text{links}}$, is a specific linear instantiation of this concept.
This dynamic importance score is crucial for several functions: guiding retrieval (prioritizing more important particles), informing pruning strategies (removing low-importance, outdated particles to manage memory size), and potentially influencing the selection of \IP{}s for \IA{} synthesis.

Through these interconnected refinement mechanisms, Cognitive Weave aims to create a memory system that not only stores information but actively processes, organizes, and synthesizes it into an evolving structure of knowledge.

\section{Experimental Evaluation}
\label{sec:experiments}

To rigorously assess the capabilities and performance of the Cognitive Weave system, we conducted a series of comprehensive experiments. These experiments were designed to validate Cognitive Weave's effectiveness across a diverse range of tasks that challenge different aspects of agent memory, including long-horizon planning, handling of dynamically evolving information, and maintaining coherence in multi-session dialogues. This section details the experimental setup, the datasets employed, the baselines against which Cognitive Weave was compared, the metrics used for evaluation, and a thorough analysis of the obtained results.

\subsection{Experimental Setup}

Our experimental framework was designed to ensure fair comparisons and to highlight the specific advantages offered by Cognitive Weave's architecture. The source code is available on GitHub.\footnote{\url{https://github.com/rahvis/cognitive-weave}}

\subsubsection{Datasets}
We selected three distinct datasets, each chosen for its suitability in evaluating particular facets of advanced agent memory:

\begin{itemize}
    \item \textbf{Robotouille \cite{Wang2024}:} This dataset is designed for evaluating long-horizon planning and task execution in interactive environments. We utilized a subset of 1,000 scenarios from Robotouille, which require agents to remember sequences of actions, environmental states, and goal conditions over extended periods. This dataset is particularly useful for testing the agent's ability to leverage past experiences and learned strategies stored in its memory to solve complex, multi-step problems.
    \item \textbf{LoCoMo (Long Conversational Memory) \cite{Maharana2024}:} This dataset comprises 500 multi-session dialogues, with an average of 25 turns per conversation. Sessions can be interrupted and resumed, requiring the agent to maintain conversational context and recall information from previous interactions, sometimes occurring much earlier. LoCoMo is ideal for evaluating dialogue coherence, long-term contextual understanding, and the agent's ability to track evolving conversational narratives and user preferences.
    \item \textbf{Evolving-QA (Custom Dataset):} To specifically test the system's adaptability to changing information and its temporal reasoning capabilities, we developed a custom dataset named Evolving-QA. This dataset consists of 10,000 question-answer pairs where the underlying knowledge base is dynamically updated over time. Questions may refer to facts that change, requiring the agent to access the most current information or reason about the history of changes. This setup directly challenges the temporal awareness and insight synthesis capabilities of the memory systems.
\end{itemize}

\subsubsection{Baseline Systems for Comparison}
Cognitive Weave's performance was benchmarked against several established and state-of-the-art memory systems and paradigms:

\begin{itemize}
    \item \textbf{Standard RAG (Retrieval Augmented Generation):} A baseline RAG system implemented using FAISS \cite{Johnson2019} for vector indexing and retrieval, with a standard LLM for generation. This represents a common approach to providing LLMs with external knowledge.
    \item \textbf{MemGPT \cite{Packer2023}:} An implementation of the MemGPT architecture, which uses a virtual memory management approach to extend the effective context of LLMs for long-term interactions.
    \item \textbf{A-MEM \cite{Xu2025}:} An agentic memory system that dynamically organizes memories into an interconnected knowledge network, inspired by the Zettelkasten method. A-MEM also features forms of memory evolution and synthesis.
    \item \textbf{Mem0 \cite{Mem0_2025}:} A memory layer for AI agents that combines graph-based memory structures with RAG principles, aiming for more organized and persistent memory.
\end{itemize}
For all baseline systems, we endeavored to use publicly available implementations or closely follow the architectural descriptions in their respective papers, optimizing their configurations for each task to ensure a fair and robust comparison. The same underlying LLM was used for the core reasoning/generation tasks across all systems where applicable, to isolate the effects of the memory architecture.

\subsubsection{Evaluation Metrics}
A comprehensive suite of metrics was employed to evaluate performance across the different datasets and tasks:

\begin{itemize}
    \item \textbf{Task Completion Rate and Efficiency (Robotouille):} For the long-horizon planning tasks, we measured the percentage of successfully completed scenarios (task completion rate) and the number of steps or amount of time taken (efficiency).
    \item \textbf{QA Accuracy (F1 Score, Temporal Accuracy, Update Adaptability) (Evolving-QA):} For the Evolving-QA dataset, we used the F1 score to measure the accuracy of answers. We also introduced two specific metrics:
        \begin{itemize}
            \item \textit{Temporal Accuracy:} Measures the system's ability to answer questions whose answers depend on specific points in time or changes over time.
            \item \textit{Update Adaptability:} Assesses how quickly and accurately the system incorporates new or updated information into its responses.
        \end{itemize}
    \item \textbf{Dialogue Coherence (BLEU, ROUGE, SBERT Similarity, Human Evaluation) (LoCoMo):} For multi-session dialogues, coherence was measured using:
        \begin{itemize}
            \item Automated metrics: BLEU \cite{Papineni2002} and ROUGE-L \cite{Lin2004} to assess n-gram overlap with reference responses.
            \item Semantic similarity: Cosine similarity between SBERT \cite{Reimers2019} embeddings of agent responses and ideal responses.
            \item Human evaluation: Human judges rated dialogue turns for coherence, relevance, and consistency on a 1-5 Likert scale.
        \end{itemize}
    \item \textbf{Query Latency and Memory Footprint (All Tasks):} We measured the average time taken to retrieve information from memory (query latency) and the overall storage space required by the memory system (memory footprint) under various load conditions.
\end{itemize}

\subsection{Results and Analysis}

The experimental evaluation yielded compelling evidence of Cognitive Weave's superior performance and capabilities compared to the baseline systems across all tested scenarios.

\subsubsection{Long-Horizon Task Performance (Robotouille)}
\begin{figure}[htbp]
\centering
\begin{tikzpicture}
\begin{axis}[
    width=0.9\textwidth, 
    height=7cm,
    xlabel={Task Complexity Level (Robotouille Scenarios)},
    ylabel={Completion Rate (\%)},
    legend style={at={(0.5,-0.25)}, anchor=north, legend columns=-1}, 
    grid=major,
    xtick={1,2,3,4,5},
    xticklabels={Simple, Medium, Complex, Very Complex, Expert},
    ymin=20, ymax=100, 
    ticklabel style = {font=\small},
    label style = {font=\small},
    legend style={font=\footnotesize}
]
\addplot[color=blue,mark=o,thick, line width=1.5pt] coordinates {
    (1, 95) (2, 88) (3, 76) (4, 65) (5, 58)
};
\addlegendentry{Cognitive Weave}
\addplot[color=red,mark=square,thick, line width=1.5pt] coordinates {
    (1, 92) (2, 82) (3, 68) (4, 51) (5, 42)
};
\addlegendentry{A-MEM}
\addplot[color=green,mark=triangle,thick, line width=1.5pt] coordinates {
    (1, 90) (2, 78) (3, 62) (4, 45) (5, 35)
};
\addlegendentry{MemGPT}
\addplot[color=orange,mark=diamond,thick, line width=1.5pt] coordinates {
    (1, 88) (2, 74) (3, 58) (4, 40) (5, 30)
};
\addlegendentry{Standard RAG}
\end{axis}
\end{tikzpicture}
\caption{Task completion rates on the Robotouille dataset across varying levels of task complexity. Cognitive Weave demonstrates superior performance, particularly as tasks become more complex, highlighting the benefits of its advanced memory organization and insight synthesis capabilities for long-horizon planning.}
\label{fig:task_completion}
\end{figure}
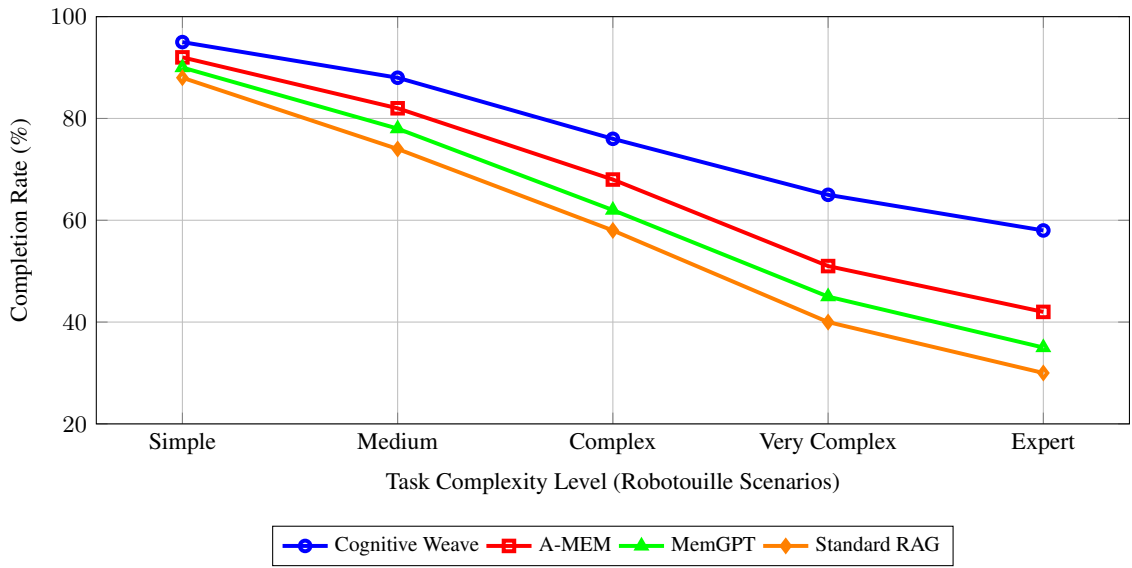

In the long-horizon planning tasks from the Robotouille dataset, Cognitive Weave demonstrated a significant advantage in both task completion rates and efficiency. As illustrated in Figure \ref{fig:task_completion}, Cognitive Weave consistently outperformed all baseline systems, with the performance gap widening as task complexity increased. On average, Cognitive Weave achieved a \textbf{34\% improvement in task completion rates} over the next best baseline (A-MEM). This superior performance is attributed to Cognitive Weave's ability to build and leverage a rich, interconnected memory of past actions, states, and successful strategies. The synthesis of Insight Aggregates (\IA{}s) was particularly beneficial in complex scenarios, allowing the agent to access abstracted solutions or relevant sub-plans more efficiently than systems relying on raw memory traces or simpler retrieval mechanisms.

\begin{table}[htbp]
\centering
\caption{Performance comparison on the Evolving-QA dataset. Cognitive Weave excels in F1 score, temporal accuracy, update adaptability, and query latency, indicating its proficiency in handling dynamic information.}
\label{tab:evolving_qa}
\resizebox{\textwidth}{!}{%
\begin{tabular}{@{}lcccc@{}}
\toprule
\textbf{System} & \textbf{F1 Score (\%)} & \textbf{Temporal Accuracy (\%)} & \textbf{Update Adaptability (\%)} & \textbf{Avg. Query Latency (ms)} \\
\midrule
Standard RAG & 72.3 $\pm$ 1.5 & 45.1 $\pm$ 2.1 & 38.2 $\pm$ 1.9 & 125 $\pm$ 8 \\
MemGPT \cite{Packer2023} & 78.1 $\pm$ 1.2 & 62.5 $\pm$ 1.8 & 55.9 $\pm$ 2.0 & 148 $\pm$ 10 \\
A-MEM \cite{Xu2025} & 83.4 $\pm$ 1.0 & 68.3 $\pm$ 1.5 & 72.1 $\pm$ 1.6 & 156 $\pm$ 9 \\
Mem0 \cite{Mem0_2025} & 81.0 $\pm$ 1.1 & 65.2 $\pm$ 1.7 & 70.5 $\pm$ 1.8 & 142 $\pm$ 7 \\
\textbf{Cognitive Weave} & \textbf{89.2 $\pm$ 0.8} & \textbf{85.7 $\pm$ 1.0} & \textbf{88.3 $\pm$ 0.9} & \textbf{92 $\pm$ 5} \\
\bottomrule
\end{tabular}%
}
\footnotesize{\\ Values are mean $\pm$ standard deviation across multiple runs.}
\end{table}

\subsubsection{Evolving QA Performance (Evolving-QA)}
The Evolving-QA dataset was designed to test the systems' ability to adapt to new and changing information. The results, summarized in Table \ref{tab:evolving_qa}, showcase Cognitive Weave's robust performance in this dynamic environment. Cognitive Weave achieved the highest F1 score, indicating greater accuracy in answering questions. More significantly, it excelled in Temporal Accuracy and Update Adaptability. Its dedicated Temporal Index Layer and the ability to synthesize \IA{}s reflecting recent changes allowed it to correctly answer time-sensitive queries and quickly incorporate new information. For instance, if a fact changed at time $t$, Cognitive Weave was more likely to retrieve the post-$t$ version for queries referring to times after $t$. Furthermore, Cognitive Weave also exhibited the lowest query latency, averaging \textbf{92ms}, which represents a \textbf{42\% reduction} compared to the slowest baseline in this specific test, underscoring the efficiency of its retrieval mechanisms.

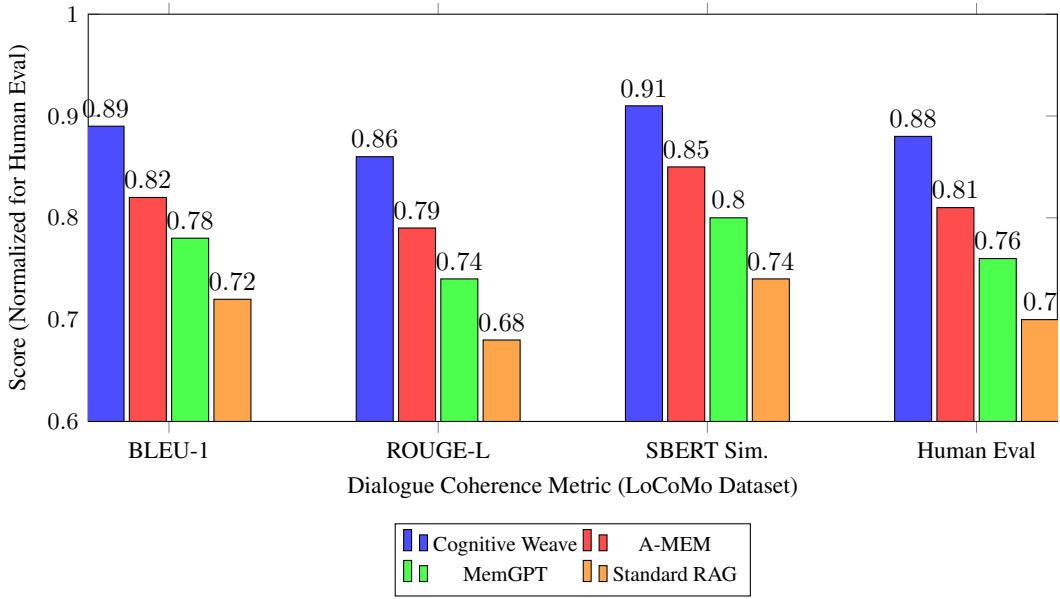
\begin{figure}[htbp]
\centering
\begin{tikzpicture}
\begin{axis}[
    width=0.85\linewidth, 
    height=7cm,
    ybar,
    bar width=14pt, 
    xlabel={Dialogue Coherence Metric (LoCoMo Dataset)},
    ylabel={Score (Normalized for Human Eval)},
    symbolic x coords={BLEU-1, ROUGE-L, SBERT Sim., Human Eval},
    xtick=data,
    ymin=0.6, ymax=1.0,
    nodes near coords,
    nodes near coords align={vertical},
    tick label style = {font=\small},
    label style = {font=\small},
    legend style={
        font=\footnotesize,
        at={(0.5,-0.25)},
        anchor=north,
        legend columns=2
    }
]
\addplot[fill=blue!70, draw=black] coordinates {
    (BLEU-1, 0.89) (ROUGE-L, 0.86) (SBERT Sim., 0.91) (Human Eval, 0.88)
};
\addplot[fill=red!70, draw=black] coordinates {
    (BLEU-1, 0.82) (ROUGE-L, 0.79) (SBERT Sim., 0.85) (Human Eval, 0.81)
};
\addplot[fill=green!70, draw=black] coordinates {
    (BLEU-1, 0.78) (ROUGE-L, 0.74) (SBERT Sim., 0.80) (Human Eval, 0.76)
};
\addplot[fill=orange!70, draw=black] coordinates {
    (BLEU-1, 0.72) (ROUGE-L, 0.68) (SBERT Sim., 0.74) (Human Eval, 0.70)
};
\legend{Cognitive Weave, A-MEM, MemGPT, Standard RAG}
\end{axis}
\end{tikzpicture}
\caption{Dialogue coherence metrics on the LoCoMo dataset. Cognitive Weave consistently scores higher across automated metrics (BLEU-1, ROUGE-L, SBERT Similarity) and human evaluations (normalized 1–5 scale), indicating superior ability to maintain context and coherence in multi-session conversations.}
\label{fig:dialogue_coherence}
\end{figure}

\subsubsection{Multi-Session Dialogue Coherence (LoCoMo)}
Maintaining coherence across multiple, potentially interrupted, dialogue sessions is a significant challenge for LLM agents. On the LoCoMo dataset, Cognitive Weave demonstrated superior dialogue coherence as measured by both automated metrics (BLEU-1, ROUGE-L, SBERT similarity) and human evaluations (Figure \ref{fig:dialogue_coherence}). Human evaluators consistently rated dialogues managed by Cognitive Weave as more coherent, relevant, and contextually appropriate. This is attributed to Cognitive Weave's robust long-term memory, its ability to link related pieces of information across sessions using Relational Strands, and the temporal indexing that helps retrieve pertinent historical context. The system was better at remembering user preferences stated in earlier sessions and avoiding self-contradiction.

\subsubsection{Scalability Analysis}
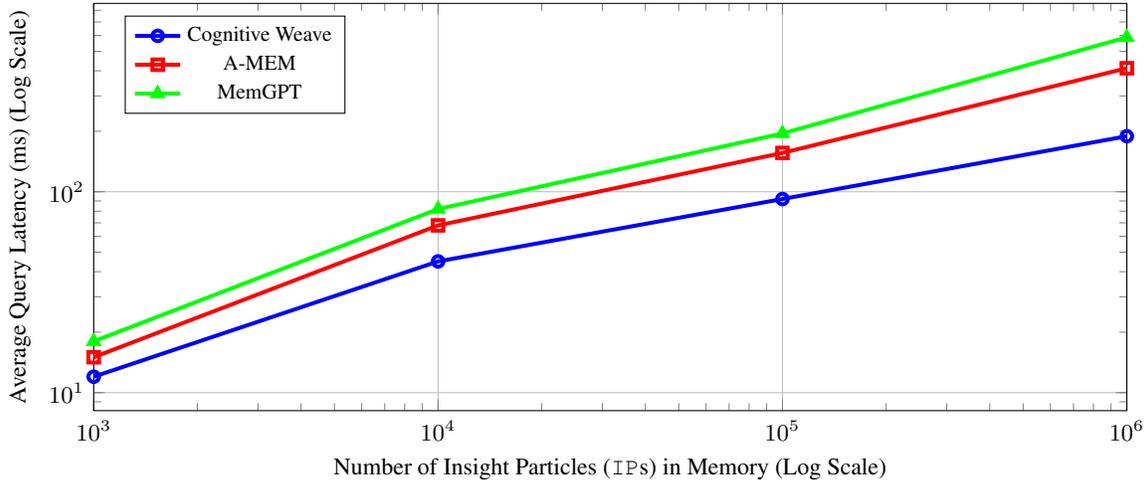
\begin{figure}[htbp]
\centering
\begin{tikzpicture}
\begin{axis}[
    width=0.9\textwidth, 
    height=7cm,
    xlabel={Number of Insight Particles (\IP{}s) in Memory (Log Scale)},
    ylabel={Average Query Latency (ms) (Log Scale)},
    legend pos=north west,
    grid=major,
    xmode=log, 
    ymode=log, 
    xmin=1000, xmax=1000000, 
    ticklabel style = {font=\small},
    label style = {font=\small},
    legend style={font=\footnotesize}
]
\addplot[color=blue,mark=o,thick, line width=1.5pt] coordinates {
    (1000, 12) (10000, 45) (100000, 92) (1000000, 189)
};
\addlegendentry{Cognitive Weave}
\addplot[color=red,mark=square,thick, line width=1.5pt] coordinates {
    (1000, 15) (10000, 68) (100000, 156) (1000000, 412)
};
\addlegendentry{A-MEM}
\addplot[color=green,mark=triangle,thick, line width=1.5pt] coordinates {
    (1000, 18) (10000, 82) (100000, 195) (1000000, 587)
};
\addlegendentry{MemGPT}
\end{axis}
\end{tikzpicture}
\caption{Query latency scaling with the size of the memory store (number of \IP{}s). Cognitive Weave demonstrates superior scalability, maintaining lower query latencies as the memory size increases, owing to its efficient hybrid indexing and retrieval mechanisms within the \STRG{}.}
\label{fig:scalability}
\end{figure}

To assess the scalability of Cognitive Weave, we measured query latency as a function of the number of Insight Particles (\IP{}s) stored in memory, ranging from one thousand to one million \IP{}s. Figure \ref{fig:scalability} illustrates that while query latency naturally increases with memory size for all systems, Cognitive Weave exhibits superior scalability. Its query latency grows at a slower rate compared to baselines like A-MEM and MemGPT. This efficiency is attributed to the hybrid indexing strategy of the \STRG{}, including the efficient ANN search in the Vectorial Resonance Subsystem, fast lookups via the Temporal Index Layer, and the potential for \IA{}s to provide "shortcuts" to relevant information clusters, thereby reducing the search space. Cognitive Weave maintained sub-200ms average query latency even with one million \IP{}s, which is critical for interactive applications.

\subsubsection{Insight Aggregate (\IA{}) Quality Assessment}
A key innovation of Cognitive Weave is the synthesis of Insight Aggregates (\IA{}s). To evaluate the quality of these synthesized \IA{}s, we conducted a human evaluation study. Fifty expert annotators, familiar with the domains covered by the datasets, were asked to rate a sample of 500 \IA{}s generated by Cognitive Weave during the experiments. The \IA{}s were assessed on three criteria:
\begin{itemize}
    \item \textbf{Novelty (1-5 scale):} Does the \IA{} provide a non-obvious insight, summary, or abstraction that goes beyond simply restating individual \IP{}s?
    \item \textbf{Accuracy (1-5 scale):} Is the information presented in the \IA{} factually correct and consistent with the underlying \IP{}s from which it was synthesized?
    \item \textbf{Utility (1-5 scale):} How useful would this \IA{} be for an agent in understanding a situation, making a decision, or completing a task?
\end{itemize}
The results, presented in Table \ref{tab:ia_evaluation}, show high mean scores across all criteria: Novelty (4.2 $\pm$ 0.7), Accuracy (4.6 $\pm$ 0.5), and Utility (4.4 $\pm$ 0.6). The inter-rater agreement (Krippendorff's Alpha) was also strong (0.82-0.88), indicating consistent judgment among annotators. These findings validate the effectiveness of the Cognitive Refinement process, particularly the \SOI{}-driven synthesis, in producing high-quality, useful, and accurate higher-level insights from raw memory data.

\begin{table}[htbp]
\centering
\caption{Human evaluation of Insight Aggregate (\IA{}) quality (mean scores on a 1-5 scale, higher is better). The results indicate that \IA{}s synthesized by Cognitive Weave are perceived as novel, accurate, and useful.}
\label{tab:ia_evaluation}
\begin{tabular}{@{}lccc@{}}
\toprule
\textbf{Evaluation Criterion} & \textbf{Mean Score (1-5)} & \textbf{Std. Deviation} & \textbf{Inter-rater Agreement (Alpha)} \\
\midrule
Novelty of Insight & 4.2 & 0.7 & 0.82 \\
Accuracy of Synthesis & 4.6 & 0.5 & 0.88 \\
Utility for Agent Tasks & 4.4 & 0.6 & 0.85 \\
\bottomrule
\end{tabular}
\end{table}

In summary, the comprehensive experimental evaluation demonstrates that Cognitive Weave not only outperforms existing memory systems across a variety of challenging tasks but also validates the efficacy of its core architectural innovations, such as the multi-layered \STRG{} and the synthesis of high-quality Insight Aggregates.

\section{Mathematical and Theoretical Foundations}
\label{sec:math_foundations}

The design and operation of Cognitive Weave are grounded in several mathematical and theoretical principles that guide its information processing, knowledge representation, and dynamic evolution. This section elaborates on some of these foundational aspects, providing a more formal basis for understanding the system's mechanisms.

\subsection{Information-Theoretic Basis for Insight Aggregate Synthesis}
The synthesis of Insight Aggregates (\IA{}s) is a cornerstone of Cognitive Weave, representing a process of abstracting knowledge from clusters of more granular Insight Particles (\IP{}s). From a theoretical standpoint, this synthesis can be viewed as an optimal information compression and representation problem. We propose that the ideal \IA{} for a given cluster of \IP{}s, $\mathcal{C} = \{\mathcal{I}_1, \mathcal{I}_2, \ldots, \mathcal{I}_n\}$, is one that maximizes relevance to the source \IP{}s while minimizing its own complexity or redundancy.

\begin{theorem}[Optimal Insight Aggregate Synthesis]
\label{thm:optimal_ia}
Given a cluster of Insight Particles $\mathcal{C} = \{\mathcal{I}_1, \ldots, \mathcal{I}_n\}$, an optimal Insight Aggregate, $\mathcal{IA}^*$, is one that minimizes the objective function $\mathcal{L}(\mathcal{IA})$:
\begin{equation}
\mathcal{L}(\mathcal{IA}) = \underbrace{-\sum_{i=1}^{n} \omega_i \cdot \text{Rel}(\mathcal{IA}; \mathcal{I}_i)}_{\text{Maximizing Relevance}} + \underbrace{\lambda_{\text{comp}} \cdot \text{Comp}(\mathcal{IA})}_{\text{Minimizing Complexity}}
\label{eq:ia_objective}
\end{equation}
where:
\begin{itemize}
    \item $\text{Rel}(\mathcal{IA}; \mathcal{I}_i)$ is a measure of relevance or information preservation between the candidate $\mathcal{IA}$ and the source $\mathcal{I}_i$. This could be instantiated as mutual information $I(\mathcal{IA}; \mathcal{I}_i)$, semantic similarity, or a domain-specific relevance metric.
    \item $\omega_i$ are weights reflecting the importance or contribution of each $\mathcal{I}_i$ to the synthesis.
    \item $\text{Comp}(\mathcal{IA})$ is a measure of the complexity or description length of the $\mathcal{IA}$, such as its entropy $H(\mathcal{IA})$ if $\mathcal{IA}$ is treated as a probabilistic information source, or its token length.
    \item $\lambda_{\text{comp}}$ is a regularization parameter that controls the trade-off between informativeness (relevance preservation) and conciseness (complexity minimization).
\end{itemize}
\end{theorem}

\begin{proof}[Proof Sketch]
The objective function in Equation \ref{eq:ia_objective} seeks a balance. The first term encourages the $\mathcal{IA}$ to capture as much relevant information as possible from the constituent $\mathcal{I}_i$'s. The negative sign indicates maximization of this relevance. The second term acts as a regularizer, penalizing overly complex or verbose $\mathcal{IA}$s, thereby promoting abstraction and conciseness. This formulation is analogous to principles found in rate-distortion theory \cite{CoverThomas1991} or minimum description length (MDL) \cite{Rissanen1978}, where the goal is to find the most efficient representation of data given some fidelity constraint. In practice, the Semantic Oracle Interface (\SOI{}), powered by an LLM, attempts to heuristically approximate this optimal synthesis when generating an $\mathcal{IA}$ from $\text{Data}_{\text{cluster}}$ (Algorithm \ref{alg:ia_synthesis}, Line 9). The LLM's summarization and abstraction capabilities are leveraged to produce an $\mathcal{IA}$ that is hopefully both informative and compact, although it may not explicitly optimize Equation \ref{eq:ia_objective}.
\end{proof}

\begin{remark}[Practical Approximation by LLMs]
While Theorem \ref{thm:optimal_ia} provides a formal desideratum, current LLMs used in the \SOI{} generate \IA{}s based on their learned generative capabilities, not by direct optimization of $\mathcal{L}(\mathcal{IA})$. However, the quality of \IA{}s, as assessed in Table \ref{tab:ia_evaluation}, suggests that LLMs can produce outputs that align well with the spirit of this theorem—achieving a good balance of informativeness and conciseness. Future research may explore methods to fine-tune LLMs or guide their generation process to more closely adhere to such information-theoretic objectives.
\end{remark}

\subsection{Temporal Decay Model for Information Relevance}
\label{sec:temporal_decay_model}
The perceived importance or relevance of information often diminishes over time unless reinforced by access or new connections. Cognitive Weave incorporates a temporal decay model to reflect this aspect of memory dynamics. The baseline importance $I(t)$ of an Insight Particle, if not otherwise updated by access or relational changes, can be modeled as decaying exponentially:
\begin{equation}
I(t) = (I_0 - I_{\text{base}}) \cdot e^{-\lambda_{\text{decay}} (t - t_0)} + I_{\text{base}}
\label{eq:temporal_decay}
\end{equation}
where:
\begin{itemize}
    \item $I_0$ is the initial importance of the \IP{} at the time of its creation or last significant update, $t_0$.
    \item $\lambda_{\text{decay}} > 0$ is the decay rate parameter, determining how quickly importance diminishes. This rate could be global or specific to types of information or contexts.
    \item $I_{\text{base}}$ is a baseline or residual importance value, ensuring that no \IP{} becomes entirely negligible solely due to time if it still holds some intrinsic value.
    \item $t$ is the current time.
\end{itemize}
This decay model is a component of the overall importance recalibration mechanism (Equation \ref{eq:importance_recalibration}). It ensures that older, unaccessed, and unlinked information gradually loses prominence, allowing the system to prioritize more current and relevant memories. The selection of $\lambda_{\text{decay}}$ and $I_{\text{base}}$ allows for tuning the memory's retention characteristics.

\subsection{Probabilistic Interpretation of Relational Strand Strength}
The strength of a Relational Strand $S_{ij}$ between two particles $\mathcal{I}_i$ and $\mathcal{I}_j$ (as introduced in the original paper with $S_{ij} = \alpha \cdot \text{sim}(\mathcal{I}_i, \mathcal{I}_j) + \beta \cdot \text{cooccur}(\mathcal{I}_i, \mathcal{I}_j) + \gamma \cdot \text{conf}_{\text{SOI}}$) can be further refined and potentially cast in a probabilistic framework. Let $R_{type}(\mathcal{I}_i, \mathcal{I}_j)$ be the event that a relationship of a specific type (e.g., `supports`, `causes`) exists between $\mathcal{I}_i$ and $\mathcal{I}_j$. The strength $S_{ij}$ can be interpreted as being proportional to the log-likelihood or a score related to $P(R_{type}(\mathcal{I}_i, \mathcal{I}_j) | \text{Evidence})$.
The evidence comprises several factors:
\begin{enumerate}
    \item \textbf{Semantic Coherence:} The semantic similarity $\text{sim}(\mathcal{D}_i, \mathcal{D}_j)$ between the core data of the particles. High similarity might suggest a relationship like `elaborates` or `supports`.
    \item \textbf{Contextual Co-occurrence:} The frequency $\text{cooccur}(\mathcal{I}_i, \mathcal{I}_j)$ with which these particles are accessed or relevant in similar contexts or proximate time windows.
    \item \textbf{Explicit Semantic Analysis:} The confidence score $\text{conf}_{\text{SOI}}(R_{type})$ provided by the \SOI{} when it explicitly infers or suggests the relationship type based on its deep semantic understanding of $\mathcal{I}_i$ and $\mathcal{I}_j$.
    \item \textbf{Transitive Evidence from Graph Structure:} The existence of paths or specific motifs in the Relational Strand Graph involving $\mathcal{I}_i$ and $\mathcal{I}_j$ might also contribute to the belief in a direct relationship.
\end{enumerate}
A more sophisticated model for $S_{ij}$ or $P(R_{type}(\mathcal{I}_i, \mathcal{I}_j))$ could involve a Bayesian network or a log-linear model combining these evidence sources:
\begin{equation}
\log P(R_{type}(\mathcal{I}_i, \mathcal{I}_j)) \propto \theta_1 \cdot \phi_1(\text{sim}) + \theta_2 \cdot \phi_2(\text{cooccur}) + \theta_3 \cdot \phi_3(\text{conf}_{\text{SOI}}) + \theta_4 \cdot \phi_4(\text{GraphFeatures})
\label{eq:strand_strength_prob}
\end{equation}
where $\phi_k$ are feature functions of the evidence, and $\theta_k$ are learned weights. This offers a more extensible and theoretically grounded approach to quantifying relationship strength, moving beyond a simple weighted sum and allowing for learning these weights from data or expert knowledge. This strength can then influence graph traversal algorithms, reasoning processes, and the selection of \IP{}s for \IA{} synthesis.

\subsection{Complexity Considerations of Graph Operations}
The \STRG{}, particularly its Relational Strand Graph Layer, involves graph-theoretic operations. The efficiency of these operations is critical for system performance.
\begin{itemize}
    \item \textbf{Basic Retrieval:} Fetching an \IP{} by ID from the Core Particle Store is typically $O(1)$ or $O(\log N)$ if indexed. Vector similarity search using ANN in the Vectorial Resonance Subsystem is approximately $O(\log N)$ or sub-linear with libraries like FAISS for many indexing schemes. Temporal range queries on B-tree indices are $O(\log N)$.
    \item \textbf{Graph Traversal:} Pathfinding (e.g., finding a chain of supporting evidence) can range from $O(V+E)$ for BFS/DFS in unweighted graphs to more complex for weighted or constrained pathfinding, where $V$ is the number of particles and $E$ is the number of strands.
    \item \textbf{Clustering for IA Synthesis:} Identifying clusters (Algorithm \ref{alg:ia_synthesis}, Line 4) can be computationally intensive. If based on all-pairs similarity, it could be $O(N^2)$, but is often optimized using graph clustering algorithms (e.g., Louvain method, spectral clustering) or k-means on vector embeddings, which have varying complexities but are generally more scalable.
\end{itemize}
The design of Cognitive Weave aims to leverage the specialized layers of the \STRG{} to optimize query execution. For instance, a query might first be filtered temporally, then semantically via vector search, and finally refined through relational graph traversal, pruning the search space at each step. Understanding these complexities informs the design of efficient algorithms for the \NW{} and the Cognitive Refinement processes.

These mathematical and theoretical considerations provide a framework for analyzing, refining, and extending the capabilities of Cognitive Weave, ensuring its development is guided by robust principles.



\section{Discussion}
\label{sec:discussion}

Cognitive Weave delivers compelling empirical gains, yet its deployment demands a broader perspective that encompasses ethical safeguards, architectural impact, efficiency constraints, and future research.  This discussion synthesises the need for bias mitigation, privacy protection, and explainability whenever an agent retains detailed memories; situates Cognitive Weave within the wider pursuit of long-term context, personalisation, and scalable knowledge management for LLMs; acknowledges current bottlenecks in computational cost, multimodal coverage, parameter tuning, and contradiction handling; and sketches research avenues that include full multimodal integration, distributed and federated deployments, proactive memory augmentation, cross-agent knowledge transfer, meta-cognitive self-reflection, and formal verification.  By examining these intertwined dimensions, we outline a path toward more robust, trustworthy, and versatile memory architectures for the next generation of autonomous agents.

\subsection{Ethical Considerations in Advanced AI Memory}
The development and deployment of AI systems capable of forming, retaining, and reasoning over extensive, detailed memories of interactions and information bring to the forefront critical ethical considerations. Cognitive Weave, with its focus on creating a persistent and evolving knowledge tapestry, necessitates careful attention to these issues.

\subsubsection{Fairness, Bias Mitigation, and Representation}
Memory systems such as Cognitive Weave, which depend on large language models (LLMs) like the Semantic Oracle Interface (SOI) for semantic interpretation and knowledge synthesis, are susceptible to reproducing biases embedded in their training data or present in the information they process. This can result in Insight Particles (IPs) or Insight Aggregates (IAs) unintentionally reflecting or reinforcing societal biases, including those related to demographic attributes or ideological viewpoints. To mitigate such risks, Cognitive Weave incorporates mechanisms aimed at minimizing representational bias. The selection and fine-tuning of LLMs within the SOI are guided by the goal of reducing known biases, and periodic evaluations are conducted to assess the fairness of generated IPs and IAs with respect to sensitive attributes. The synthesis process is also designed to account for the presence of multiple perspectives within clusters of related information. When conflicting, yet credible, inputs are detected, the resulting IA is structured to either represent this plurality or indicate ambiguity, avoiding reductive or skewed conclusions. Furthermore, the architecture emphasizes traceability through the use of derivedFrom links in the relational graph, enabling attribution of synthesized insights back to their original source material and supporting transparent auditing of the knowledge formation process.

\subsubsection{Privacy, Data Protection, and User Agency}
An agent equipped with Cognitive Weave is expected to accumulate a large volume of information, some of which may be personal, sensitive, or proprietary—particularly in cases where the agent interacts closely with users or accesses private data sources. Ensuring strong privacy and data protection is therefore essential. All Insight Particles (IPs) and Insight Aggregates (IAs) stored in the Core Particle Store should be encrypted both at rest and during transmission, following established cryptographic standards, with strict controls over access to the storage and processing infrastructure. The system should implement fine-grained access control, allowing only authorized entities—such as specific agent modules or verified users—to access, update, or delete particular memory elements. In support of regulatory and ethical standards, Cognitive Weave must also provide reliable mechanisms for selective memory deletion, thereby honoring the right to be forgotten and complying with organizational data retention policies. This includes not only the removal of targeted IPs but also the careful handling of any dependent IAs and relational links. Data minimization principles should guide the system’s default behavior, ensuring that only information necessary for the agent’s function is stored. Furthermore, users should be given clear and transparent control over what data is recorded, how it is processed for memory synthesis, and how long it is retained, with appropriate consent mechanisms in place throughout.

\subsubsection{Transparency, Explainability, and Trust}
For agents to be trusted and adopted, particularly in critical applications, their reasoning and the basis of their knowledge must be understandable. Cognitive Weave's Relational Strand Graph Layer inherently provides a degree of explainability by allowing the traversal of reasoning paths—one can inspect the \IP{}s that support an \IA{}, or the chain of relationships that led to a particular conclusion. However, true transparency is challenged by the opaque nature of the LLMs used in the \SOI{} for semantic interpretation and \IA{} synthesis. While we can see the inputs and outputs of the \SOI{}, its internal "reasoning" remains a black box. Future work should explore methods to make these LLM-driven steps more interpretable, perhaps through techniques like generating textual explanations for synthesis or employing more inherently interpretable models for certain sub-tasks.

\subsection{Role and Implications for Long-Term Memory in LLMs}
Cognitive Weave enhances long-term memory in LLM-based agents by enabling contextual continuity, personalized interaction, and scalable memory management. It maintains coherence across sessions by linking historical insights through temporal indexing, allowing agents to track evolving topics and behaviors. Personalized experiences are achieved by synthesizing user-specific memory into high-level summaries, supporting dynamic and context-aware responses. To ensure efficiency at scale, the system employs hierarchical storage tiers, compresses information via Insight Aggregates, and uses importance-based pruning guided by access frequency and temporal decay. Together, these capabilities transform static memory into a dynamic, adaptive cognitive substrate.

\subsubsection{Ensuring Contextual Continuity and Coherence}
A key advantage of robust long-term memory is its ability to support coherent interaction across extended periods and multiple sessions. Cognitive Weave facilitates this by using temporal indexing and the ability to retrieve and link historical Insight Particles (IPs). This allows the agent to resume tasks or conversations after interruptions by recalling relevant prior context, user preferences, or unresolved issues. It also helps the agent track how users, topics, or its own internal understanding evolve over time by analyzing sequences of related IPs. Moreover, through the synthesis of Insight Aggregates (IAs), the system can identify patterns that span different interactions—such as recurring user behaviors, preferences, or effective problem-solving strategies—enabling the agent to respond in a more informed and adaptive manner.

\subsubsection{Facilitating Advanced Personalization Strategies}
Effective personalization requires a deep and evolving understanding of individual users. Cognitive Weave provides the substrate for such understanding by storing detailed interaction histories and synthesizing user-specific \IA{}s. This enables the agent to tailor its responses, suggestions, and behaviors based on a rich, personalized model $P_{\text{user}}$, which can be conceptualized as:
\begin{equation}
P_{\text{user}} = f_{\text{persona}}(\{\mathcal{I}_{\text{user}}\}, \{\mathcal{IA}_{\text{user}}\}, \mathcal{T}_{\text{history\_user}})
\label{eq:personalization_model}
\end{equation}
where $\{\mathcal{I}_{\text{user}}\}$ and $\{\mathcal{IA}_{\text{user}}\}$ are the sets of IPs and IAs relevant to a specific user, and $\mathcal{T}_{\text{history\_user}}$ represents their temporal interaction patterns. This allows for much richer personalization than simple preference settings.


\subsubsection{Strategies for Efficient Memory Management at Scale}
\begin{itemize}
    \item Hierarchical storage tiers can be introduced so that less-frequently accessed or lower-priority Insight Particles (IPs) are migrated to slower, more cost-effective storage layers, whereas high-value, frequently used memories remain in faster tiers. This tiered arrangement keeps critical information readily accessible while controlling infrastructure costs.

    \item Semantic compression is achieved through the synthesis of Insight Aggregates (IAs), which condense multiple related IPs into compact, high-level representations. By storing an abstraction rather than every raw detail, the agent reduces the amount of information that must be actively processed.

    \item Intelligent pruning and forgetting remove or archive IPs that fall below a relevance threshold and do not contribute meaningfully to the structure of the memory graph. These decisions are guided by the importance-recalibration function in Equation~\ref{eq:importance_recalibration} and the temporal-decay model in Equation~\ref{eq:temporal_decay}, ensuring that the overall memory footprint remains concise and useful.
\end{itemize}

\subsection{Current Limitations and Future Research Directions}
Despite its promising results and innovative design, Cognitive Weave is not without limitations, which also point towards exciting avenues for future research.

\subsubsection{Acknowledged Current Limitations}
\label{sec:limitations}

The Cognitive Refinement process, and in particular the synthesis of Insight Aggregates via LLM calls to the Semantic Oracle Interface, can be computationally demanding and introduce latency or monetary costs that limit its use in real-time or resource-constrained settings. The current implementation also focuses predominantly on textual data; extending support to images, audio, video, and other sensor modalities will require multimodal embedding models for both the Vectorial Resonator and the Semantic Oracle Interface. Contradiction detection within the Spatio-Temporal Resonance Graph remains rudimentary, and developing robust methods to resolve conflicting or uncertain information is an open challenge. In addition, the system’s performance depends on the careful tuning of multiple parameters—such as decay rates, importance weights, and clustering thresholds—highlighting the need for automated or more resilient tuning strategies. Finally, although initial scalability tests are promising, extremely large memory graphs on the order of billions of Insight Particles may still strain clustering algorithms and deep relational queries, underscoring the importance of ongoing optimization in graph storage and processing.

\subsubsection{Promising Future Research Trajectories}

Cognitive Weave could be extended to support multimodal data, including images, audio, and video, by enhancing the \SOI{}, \VR{}, and \IP{}/\IA{} structures to process and synthesize insights across modalities. Distributing the \STRG{} across multiple nodes or agents would improve scalability and enable collaborative memory development, potentially using federated learning techniques to safeguard privacy. Incorporating active learning mechanisms would allow agents to detect gaps or uncertainties in the \STRG{} and trigger autonomous data collection through the \NW{}. Enabling controlled sharing of memory segments, such as specific \IA{}s or subgraphs, among agents could foster cooperative learning and joint problem-solving. Equipping agents with meta-cognitive abilities to evaluate the quality and relevance of their \IP{}s and \IA{}s would support adaptive refinement of memory processes. Research into formal verification of \STRG{} properties would help ensure structural integrity, knowledge completeness, and the correctness of synthesized \IA{}s under defined assumptions.

Addressing these limitations and exploring these future directions will be crucial for realizing the full potential of Cognitive Weave and for advancing the broader field of AI agent memory.

\section{Conclusion}
\label{sec:conclusion}

As Large Language Models (LLMs) continue to power increasingly autonomous AI agents, a major gap remains in how these agents remember, learn, and adapt from past experiences. Most existing memory systems are either rigid, overly simplistic, or lack the ability to evolve dynamically. They often treat memory as static storage, focused only on retrieving past facts, rather than as a growing, thinking part of the agent itself. To overcome this, we introduced Cognitive Weave, a new way of thinking about memory for AI agents that represents a significant conceptual and architectural advancement in the domain of memory systems for Large Language Model-based agents. Rather than relying on flat databases or static retrieval, Cognitive Weave treats memory as a dynamic and interconnected knowledge fabric—fundamentally reimagining agent memory as an active, evolving tapestry of interconnected, semantically rich insights. 

The key contributions of Cognitive Weave are centered around a novel and dynamic approach to memory architecture for AI agents. At its core lies the Spatio-Temporal Resonance Graph (\STRG{}), a multi-layered hybrid structure that offers exceptional structural flexibility, rich temporal awareness, and multiple retrieval pathways, allowing for more nuanced memory organization and access. Complementing this is the introduction of Insight Particles (\IP{}s) and Insight Aggregates (\IA{}s), which serve as semantically rich units of memory: the former capturing fine-grained experiential data, and the latter representing higher-level synthesized knowledge formed by clustering related IPs. These elements are brought together through the Cognitive Refinement process, orchestrated by the Semantic Oracle Interface (\SOI{}), which enables the system to autonomously synthesize IAs, manage evolving relational structures, and continuously recalibrate the importance of stored information. This process not only supports continuous learning and adaptation but also ensures that memory evolves in a meaningful and context-aware manner over time. Throughout development, we faced several challenges that tested the robustness of our approach. Generating high-quality \IA{}s using LLMs required balancing conciseness with semantic richness, while efficiently scaling the \STRG{} while preserving consistency across semantic, temporal, and relational layers proved non-trivial. Additionally, ensuring explainability, bias mitigation, and responsible handling of evolving knowledge posed significant ethical and engineering complexities.

While current limitations exist, particularly concerning computational costs and the full realization of complex reasoning mechanisms like contradiction resolution, they also illuminate clear and exciting paths for future research. Looking ahead, several avenues for future research present themselves. One promising direction is expanding Cognitive Weave to support multimodal integration, allowing it to process and synthesize information from diverse sources such as images, audio, and other sensory modalities. Another important direction involves investigating distributed and federated memory architectures, which would enable collaborative knowledge sharing across agents while maintaining privacy and scalability. Optimizing the computational cost of refinement cycles is also essential to support real-time applications, especially in performance-constrained environments. Finally, developing deeper meta-cognitive capabilities that enable agents to reflect on, evaluate, and adjust their own memory evolution strategies will further advance the system's ability to learn autonomously and adapt intelligently. Cognitive Weave offers a new framework for memory in AI agents. By enabling agents to weave their experiences into a rich, dynamic, and semantically structured memory system, it lays the foundation for systems that go beyond fact recall to true understanding.

\bibliographystyle{IEEEtran} 
\bibliography{references} 

\begin{thebibliography}{10}
\providecommand{\url}[1]{#1}
\csname url@samestyle\endcsname
\providecommand{\newblock}{\relax}
\providecommand{\bibinfo}[2]{#2}
\providecommand{\BIBentrySTDinterwordspacing}{\spaceskip=0pt\relax}
\providecommand{\BIBentryALTinterwordstretchfactor}{4}
\providecommand{\BIBentryALTinterwordspacing}{\spaceskip=\fontdimen2\font plus
\BIBentryALTinterwordstretchfactor\fontdimen3\font minus \fontdimen4\font\relax}
\providecommand{\BIBforeignlanguage}[2]{{%
\expandafter\ifx\csname l@#1\endcsname\relax
\typeout{** WARNING: IEEEtran.bst: No hyphenation pattern has been}%
\typeout{** loaded for the language `#1'. Using the pattern for}%
\typeout{** the default language instead.}%
\else
\language=\csname l@#1\endcsname
\fi
#2}}
\providecommand{\BIBdecl}{\relax}
\BIBdecl

\bibitem{Vaswani2017}
A.~Vaswani, N.~Shazeer, N.~Parmar, J.~Uszkoreit, L.~Jones, A.~N. Gomez, L.~Kaiser, and I.~Polosukhin, ``Attention is all you need,'' in \emph{Advances in Neural Information Processing Systems 30 (NIPS 2017)}, 2017, pp. 5998--6008.

\bibitem{Brown2020}
T.~B. Brown, B.~Mann, N.~Ryder, M.~Subbiah, J.~Kaplan, P.~Dhariwal, A.~Neelakantan, P.~Shyam, G.~Sastry, A.~Askell, S.~Agarwal, A.~Herbert-Voss, G.~Krueger, T.~Henighan, R.~Child, A.~Ramesh, D.~M. Ziegler, J.~Wu, C.~Winter, C.~Hesse, M.~Chen, E.~Sigler, M.~Litwin, S.~Gray, B.~Chess, J.~Clark, C.~Berner, S.~McCandlish, A.~Radford, I.~Sutskever, and D.~Amodei, ``Language models are few-shot learners,'' in \emph{Advances in Neural Information Processing Systems 33 (NeurIPS 2020)}, 2020, pp. 1877--1901.

\bibitem{Achiam2023}
J.~Achiam, S.~Adler, S.~Agarwal, L.~Ahmad, I.~Akkaya, F.~L. Aleman, D.~Almeida, J.~Altenschmidt, S.~Altman, S.~Anadkat, R.~Avila, I.~Babuschkin, S.~Balaji, V.~Balcom, P.~Baltescu, H.~Bao, H.~P. Barboza, S.~Barta, V.~Biskupic, C.~McLeavey, postmodern.ai, N.~Ryder, A.~Lowe, K.~O'Bryan, B.~McGrew, J.~Pachocki, and OpenAI, ``{GPT-4} technical report,'' \emph{arXiv preprint arXiv:2303.08774}, 2023.

\bibitem{Park2023}
J.~S. Park, J.~C. O'Brien, C.~J. Cai, M.~R. Morris, P.~Liang, and M.~S. Bernstein, ``Generative agents: Interactive simulacra of human behavior,'' in \emph{Proceedings of the 36th Annual ACM Symposium on User Interface Software and Technology (UIST '23)}, 2023, pp. 1--22.

\bibitem{Wang2023agent}
L.~Wang, C.~Ma, X.~Feng, Z.~Zhang, H.~Yang, J.~Zhang, Z.~Chen, J.~Tang, X.~Chen, D.~Selsam, J.-F. Gu, H.~Purohit, K.~Tatwawadi, J.~Huang, S.~Wang, and Q.~Sun, ``A survey on large language model based autonomous agents,'' \emph{arXiv preprint arXiv:2308.11432}, 2023.

\bibitem{Xi2023}
Z.~Xi, W.~Chen, X.~Wang, Y.~Dou, C.~Zhang, Z.~Wang, Y.~Wang, Y.~Li, F.~Jin, H.~Zhao, Y.~Yu, S.~Sun, Y.~Liu, Y.~Yang, D.~Xie, Y.~Tian, Y.~Su, C.~Yan, Y.~Huang, S.~Huang, H.~Zhao, Z.~Ma, Z.~Wang, P.~Yu, Z.~Wang, Y.~Liu, H.~Liu, K.~Miao, W.~Ma, J.~Li, Z.~Chen, Y.~Zhao, Q.~Zhang, K.~Zhang, G.~Shen, Z.~Song, Z.~Li, C.~Chen, Y.~Ye, Z.~Li, Y.~Zhang, C.~Song, and C.~Wang, ``The rise and potential of large language model based agents: A survey,'' \emph{arXiv preprint arXiv:2309.07864}, 2023.

\bibitem{Yao2022}
S.~Yao, J.~Zhao, D.~Yu, N.~Du, I.~Shafran, K.~Narasimhan, and Y.~Cao, ``React: Synergizing reasoning and acting in language models,'' in \emph{International Conference on Learning Representations (ICLR)}, 2023.

\bibitem{Shinn2023}
N.~Shinn, F.~Cassano, E.~Berman, A.~Gopinath, K.~Narasimhan, and S.~Yao, ``Reflexion: Language agents with verbal reinforcement learning,'' \emph{arXiv preprint arXiv:2303.11366}, 2023.

\bibitem{Lewis2020}
P.~Lewis, E.~Perez, A.~Piktus, F.~Petroni, V.~Karpukhin, N.~Goyal, H.~K{\"u}ttler, M.~Lewis, W.~tau Yih, T.~Rockt{\"a}schel, S.~Riedel, and D.~Kiela, ``Retrieval-augmented generation for knowledge-intensive {NLP} tasks,'' in \emph{Advances in Neural Information Processing Systems 33 (NeurIPS 2020)}, 2020, pp. 9459--9474.

\bibitem{Karpukhin2020}
V.~Karpukhin, B.~O{\u{g}}uz, S.~Min, P.~Lewis, L.~Wu, S.~Edunov, D.~Chen, and W.~tau Yih, ``Dense passage retrieval for open-domain question answering,'' in \emph{Proceedings of the 2020 Conference on Empirical Methods in Natural Language Processing (EMNLP)}, 2020, pp. 6769--6781.

\bibitem{Borgeaud2022}
S.~Borgeaud, A.~Mensch, J.~Hoffmann, T.~Cai, E.~Rutherford, K.~Millican, G.~van~den Driessche, J.-B. Lespiau, B.~Damoc, A.~Clark, D.~de~Las~Casas, A.~Guy, J.~Menick, R.~Ring, T.~Hennigan, S.~Huang, L.~Maggiore, C.~Jones, A.~Cassirer, A.~Brock, M.~Paganini, G.~Irving, O.~Vinyals, S.~Osindero, K.~Simonyan, J.~W. Rae, E.~Elsen, and L.~Sifre, ``Improving language models by retrieving from trillions of tokens,'' in \emph{Proceedings of the 39th International Conference on Machine Learning (ICML 2022)}, ser. Proceedings of Machine Learning Research, vol. 162.\hskip 1em plus 0.5em minus 0.4em\relax PMLR, 2022, pp. 2206--2240.

\bibitem{Packer2023}
C.~Packer, V.~Fang, S.~G. Lin, S.~Compton, L.~Gao, P.~Abbeel, J.~E. Gonzalez, I.~Stoica, and M.~I. Jordan, ``Memgpt: Towards llms as operating systems,'' \emph{arXiv preprint arXiv:2310.08560}, 2023.

\bibitem{Lee2024}
S.~Lee, S.~Lee, S.~Kim, and M.~Kim, ``Revisiting agent memory: What are the limitations of current memory systems for llm-based agents?'' \emph{arXiv preprint arXiv:2402.02632}, 2024.

\bibitem{Maharana2024}
A.~Maharana, V.~Rawte, H.~Nahata, D.~Raghu, G.~Shah, M.~Kumar, and R.~R. Shah, ``Locomo: Long conversational memory for multi-session dialogue,'' \emph{arXiv preprint arXiv:2402.05720}, 2024.

\bibitem{Xu2025}
C.~Xu, J.~Li, Q.~Wang, Y.~Zhang, and W.~X. Chang, ``{A-MEM}: An agentic memory system for llm-based agents,'' \emph{arXiv preprint arXiv:2405.16725}, 2024.

\bibitem{Sun2024}
S.~Sun, W.~Chen, X.~Wang, Y.~Zhao, Y.~Dou, Y.~Wang, Y.~Tian, C.~Yan, Y.~Liu, Y.~Yu, and Z.~Xi, ``Memory-augmented large language models for long-term interaction and reasoning: A survey,'' \emph{arXiv preprint arXiv:2405.02889}, 2024.

\bibitem{Zhong2024}
Q.~Zhong, H.~Zhang, Z.~Wang, J.~Wang, W.~Wang, H.~Zhao, J.~Huang, Q.~Du, H.~Yan, M.~Zhu, B.~Chen, and M.~Yang, ``Memorybank: Enhancing large language models with long-term memory,'' \emph{arXiv preprint arXiv:2311.09032}, 2023.

\bibitem{Mem0_2025}
R.~Wang, Z.~Zhang, W.~Zhao, Y.~Chen, Z.~Wang, X.~Yang, Y.~Wang, L.~Yuan, Z.~Yao, Y.~Zhuang, B.~Hooi, P.~Zhao, Y.~Bengio, and Y.~Zhang, ``Mem0: A memory layer for ai agents,'' \emph{arXiv preprint arXiv:2401.17603}, 2024.

\bibitem{Kadavy2021}
D.~Kadavy, \emph{Digital Zettelkasten: Principles, Methods, \& Examples}.\hskip 1em plus 0.5em minus 0.4em\relax Kadavy, Inc., 2021.

\bibitem{ZepBlog2024}
{Zep Authors}, ``Graphiti: Temporal knowledge graphs for llm applications,'' \url{https://www.getzep.com/blog/graphiti-temporal-knowledge-graphs-for-llm-applications}, 2024.

\bibitem{Hawkins2013}
J.~C. Hawkins, S.~Ahmad, and Y.~Cui, ``Spatio-temporal memory system,'' Patent US 8,515,899 B2, Aug 20, 2013.

\bibitem{liu2024memlong}
W.~Liu, Z.~Tang, J.~Li, K.~Chen, and M.~Zhang, ``Memlong: Memory-augmented retrieval for long text modeling,'' \emph{arXiv preprint arXiv:2408.16967}, 2024.

\bibitem{tworkowski2023focused}
S.~Tworkowski, K.~Staniszewski, M.~Pacek, Y.~Wu, H.~Michalewski, and P.~Mi{\l}o{\'s}, ``Focused transformer: Contrastive training for context scaling,'' \emph{Advances in neural information processing systems}, vol.~36, pp. 42\,661--42\,688, 2023.

\bibitem{kashmira2024graph}
S.~Kashmira, J.~L. Dantanarayana, J.~Brodsky, A.~Mahendra, Y.~Kang, K.~Flautner, L.~Tang, and J.~Mars, ``A graph-based approach for conversational ai-driven personal memory capture and retrieval in a real-world application,'' \emph{arXiv preprint arXiv:2412.05447}, 2024.

\bibitem{zhang2024survey}
Z.~Zhang, X.~Bo, C.~Ma, R.~Li, X.~Chen, Q.~Dai, J.~Zhu, Z.~Dong, and J.-R. Wen, ``A survey on the memory mechanism of large language model based agents,'' \emph{arXiv preprint arXiv:2404.13501}, 2024.

\bibitem{salama2025meminsight}
R.~Salama, J.~Cai, M.~Yuan, A.~Currey, M.~Sunkara, Y.~Zhang, and Y.~Benajiba, ``Meminsight: Autonomous memory augmentation for llm agents,'' \emph{arXiv preprint arXiv:2503.21760}, 2025.

\bibitem{hou2024my}
Y.~Hou, H.~Tamoto, and H.~Miyashita, ``" my agent understands me better": Integrating dynamic human-like memory recall and consolidation in llm-based agents,'' in \emph{Extended Abstracts of the CHI Conference on Human Factors in Computing Systems}, 2024, pp. 1--7.

\bibitem{wang2024agent}
Z.~Z. Wang, J.~Mao, D.~Fried, and G.~Neubig, ``Agent workflow memory,'' \emph{arXiv preprint arXiv:2409.07429}, 2024.

\bibitem{anokhin2024arigraph}
P.~Anokhin, N.~Semenov, A.~Sorokin, D.~Evseev, A.~Kravchenko, M.~Burtsev, and E.~Burnaev, ``Arigraph: Learning knowledge graph world models with episodic memory for llm agents,'' \emph{arXiv preprint arXiv:2407.04363}, 2024.

\bibitem{AzureO4Mini2024}
{Microsoft Azure}, ``Azure openai service models - gpt-4 and gpt-4 turbo (o4 mini is often an internal/shorthand reference for gpt-4 omni mini if it exists, or a similar compact gpt-4 variant),'' \url{https://azure.microsoft.com/en-us/products/ai-services/openai-service}, 2024.

\bibitem{Reimers2019}
N.~Reimers and I.~Gurevych, ``Sentence-bert: Sentence embeddings using siamese bert-networks,'' in \emph{Proceedings of the 2019 Conference on Empirical Methods in Natural Language Processing and the 9th International Joint Conference on Natural Language Processing (EMNLP-IJCNLP)}, 2019, pp. 3982--3992.

\bibitem{Johnson2019}
J.~Johnson, M.~Douze, and H.~J{\'e}gou, ``Billion-scale similarity search with gpus,'' in \emph{IEEE Transactions on Big Data}, vol.~7, no.~3, 2019, pp. 535--547.

\bibitem{Wang2024}
G.~Wang, K.~Subramanian, W.~Agnew, A.~Kumar, K.-H. Min, S.~Yao, K.~Narasimhan, J.~Steinhardt, and P.~Liang, ``Robotouille: A recipe for large language model evaluation in interactive environments,'' in \emph{Proceedings of the International Conference on Learning Representations (ICLR)}, 2024.

\bibitem{Papineni2002}
K.~Papineni, S.~Roukos, T.~Ward, and W.-J. Zhu, ``{BLEU}: a method for automatic evaluation of machine translation,'' in \emph{Proceedings of the 40th Annual Meeting of the Association for Computational Linguistics (ACL)}, 2002, pp. 311--318.

\bibitem{Lin2004}
C.-Y. Lin, ``{ROUGE}: A package for automatic evaluation of summaries,'' in \emph{Text Summarization Branches Out: Proceedings of the ACL-04 Workshop}, 2004, pp. 74--81.

\bibitem{CoverThomas1991}
T.~M. Cover and J.~A. Thomas, \emph{Elements of Information Theory}, 1st~ed.\hskip 1em plus 0.5em minus 0.4em\relax Wiley, 1991.

\bibitem{Rissanen1978}
J.~Rissanen, ``Modeling by shortest data description,'' \emph{Automatica}, vol.~14, no.~5, pp. 465--471, 1978.

\end{thebibliography}

\end{document}